\documentclass[lettersize,journal,compsoc]{IEEEtran}
\usepackage{amsfonts}

\usepackage[utf8]{inputenc} 
\usepackage[T1]{fontenc}    
\usepackage{hyperref}       
\usepackage{url}            
\usepackage{booktabs}       
\usepackage{amsfonts}       
\usepackage{nicefrac}       
\usepackage{microtype}      
\usepackage{xcolor}         
\usepackage{subfigure}

\usepackage{amsmath,amssymb,amsfonts}

\usepackage{graphicx}
\usepackage[inline]{enumitem}
\usepackage{amsmath}
\usepackage{multirow}
\usepackage{color}
\usepackage{subcaption}
\usepackage{makecell}

\usepackage{pifont}

\usepackage{xcolor,colortbl}

\usepackage{wrapfig}

\DeclareMathOperator{\argmin}{argmin}

\newenvironment{rcases}
  {\left.\begin{aligned}}
  {\end{aligned}\right\rbrace}
  
\title{PointAD+: Learning Hierarchical Representations for ZS 3D Anomaly Detection}

\author{Qihang Zhou,
        Shibo	He,~\IEEEmembership{Senior Member,~IEEE,} Jiangtao Yan,\\ 
        Wenchao	Meng,~\IEEEmembership{Senior Member,~IEEE,} and
        Jiming	Chen,~\IEEEmembership{Fellow,~IEEE}

\IEEEcompsocitemizethanks{

\IEEEcompsocthanksitem Qihang Zhou, Shibo He, Jiangtao Yan, Wenchao Meng, and Jiming Chen are with the State Key Laboratory of Industrial Control Technology, Zhejiang University, Hangzhou, Zhejiang, 310027, China. E-mail: \{zqhang, s18he, jtaoy, wmengzju, and cjm\}@zju.edu.cn. 
\IEEEcompsocthanksitem Corresponding author: Shibo He.
\IEEEcompsocthanksitem This work was supported by the National Natural Science Foundation Program of China under Grant U23A20326 and 62088101 Autonomous Intelligent
Unmanned Systems.

}
}
 
\def\ie{\emph{i.e.}}

\definecolor{orange}{rgb}{0,0,0}

\begin{document}

\IEEEtitleabstractindextext{%
\begin{abstract}
Zero-shot (ZS) 3D anomaly detection is a critical yet underexplored field that addresses scenarios where target 3D training samples are unavailable. This presents a challenge, as it requires the model to accurately detect 3D anomalies in previously unseen categories. In this paper, we aim to transfer CLIP's robust 2D generalization capabilities to identify 3D anomalies across unseen objects of highly diverse class semantics. We first design PointAD, which leverages point-pixel correspondence to represent 3D anomalies through their associated rendering pixel representations. This representation is referred to as implicit point representation, as it focuses solely on rendering pixel anomalies but neglects the inherent spatial relationships within point clouds. However, spatial awareness modeling is important for accurate 3D abnormality recognition. Therefore, we propose PointAD+ to further broaden the interpretation of 3D anomalies by introducing explicit point representation that is aggregated from implicit point representation, modeling abnormal point relative position. We introduce G-aggregation to involve geometry information to enable the explicit point representation
to be spatially aware. To simultaneously capture rendering and geometry abnormality, PointAD+ proposes hierarchical representation learning, incorporating implicit and explicit anomaly semantics into hierarchical text prompts: rendering prompts for the rendering layer and geometry prompts for the geometry layer. A cross-hierarchy contrastive alignment is introduced to promote the interaction between the rendering and geometry layers, facilitating mutual anomaly learning. Finally, PointAD+ integrates anomaly semantics from both layers to capture the generalized anomaly semantics. During the test, PointAD+ can integrate RGB information in a plug-and-play manner and further improve its detection performance. Extensive experiments demonstrate the superiority of PointAD+ in ZS 3D anomaly detection across unseen objects with highly diverse class semantics, achieving a holistic understanding of abnormality.
\end{abstract}

}

\maketitle

\section{Introduction}
\begin{figure}[t]
\begin{center}
\centerline{\includegraphics[width=0.9\columnwidth]{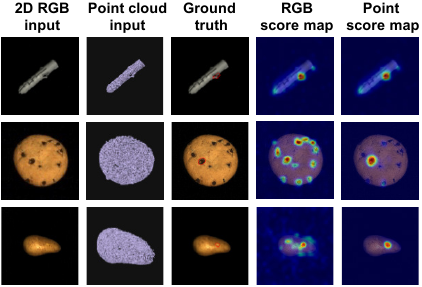}}
\caption{Motivation of zeor-shot 3D anomaly detection. \textbf{Top:} The bend in a dowel can be detected using both RGB information and point relations. \textbf{Middle and Bottom:} Challenges arise when RGB information alone misinterprets similar appearances, such as chocolates on cookies resembling hole anomalies or surface damage on a potato blending with the foreground’s color patterns. However, effective detection can be achieved by modeling point relations within point clouds.}
\label{f1: ZS results}
\end{center}
\vspace{-3.0em}
\end{figure}

Anomaly detection has garnered considerable attention from both academia and industry. It has been widely applied to diverse applications, such as industrial inspection~\cite{bergmann2019mvtec, bergmann2020uninformed, liznerski2020explainable, pang2021explainable, Zhou2023PullP, li2024clipsam, pang2021deep,roth2022towards, zuo2024clip3d, deng2022anomaly,jeong2023winclip, xie2023pushing, cao2023anomaly, gu2023anomalygpt, zhou2024pointad} and medical analysis~\cite{zhou2023anomalyclip, tian2023self, tian2021constrained}. Driven by the growing need to understand 3D objects in real-world applications, such as spatial intelligence, 3D anomaly detection has emerged as a promising field~\cite{horwitz2023back,bergmann2023anomaly,rudolph2023asymmetric,chen2023easynet,wang2023multimoda, cheng2024zeroshotpointcloudanomaly,chu2023shape,ye2025po3ad, liang2025examining}. Unlike conventional 2D anomaly detection methods that rely solely on RGB information~\cite{horwitz2023back, bergmann2023anomaly, rudolph2023asymmetric, chen2023easynet, wang2023multimoda, chu2023shape, huang2022registration, you2022unified, sun2023when}, 3D anomaly detection focuses on uncovering spatial relationships that reveal abnormal patterns. This approach is particularly effective when normal regions share similar RGB characteristics with anomalies or when defects closely mimic the appearance of an object's foreground. For instance, as shown in Figure~\ref{f1: ZS results}, chocolates (in black) on cookies closely resemble hole anomalies (also in black), causing the RGB model to mistakenly classify the chocolates as anomalies. Similarly, surface damage on a potato exhibits a color pattern similar to that of the potato's foreground, leading to misdetection by the RGB model. In contrast, 3D anomaly detection can effectively mitigate such failures by identifying abnormal spatial relationships.

Current 3D anomaly detection methods typically operate under an unsupervised setting, assuming that the target point clouds are available and entirely normal. To effectively model the normal decision boundary, these methods generally store normal point features from the training dataset as a reference memory and identify anomalies by measuring the distance between the test features and the stored ones~\cite{horwitz2023back,wang2023multimoda, chu2023shape}. However, this assumption does not hold in many situations where training samples in the target dataset are inaccessible due to privacy protection (e.g., involvement of trade secrets) or the absence of target training data (e.g., a new product that has never been seen before)~\cite{zhou2023anomalyclip}. 
The unsupervised paradigm detects class-level anomalies by memorizing or reconstructing normal point features of the target class. Hence, it is limited in its ability to capture anomalies in unseen objects. Noteably, ZS 3D anomaly detection requires the model to identify anomalies in unseen classes. While ZS anomaly detection has been explored in 2D RGB images~\cite{ming2022delving, zhou2023anomalyclip, li2024promptad}, ZS 3D anomaly detection remains an underexplored research area. This task is particularly challenging as it requires the model to capture generic anomaly semantics from a limited set of auxiliary classes, despite significant discrepancies in object semantics and the presence of unknown anomaly patterns. Recently, Vision-Language Models (VLMs) have shown strong generalization capabilities to unseen objects and have been successfully applied to various downstream tasks~\cite{radford2021learning,zhou2022learning,rao2022denseclip,sun2022dualcoop,khattak2023maple,kirillov2023segment}. One popular model, CLIP, has demonstrated impressive ZS performance in detecting 2D anomalies~\cite{ming2022delving,jeong2023winclip,zhou2023anomalyclip}. Integrating CLIP into the detection model presents a potential solution to the challenging yet unexplored domain of ZS 3D anomaly detection. It is nontrivial to directly leverage CLIP for ZS 3D anomaly detection as the knowledge gap between the 2D domain and 3D domain is significant.

\begin{figure}[t]
  \centering
    \subfigure[The schematic of PointAD.]{\includegraphics[width=1\columnwidth]{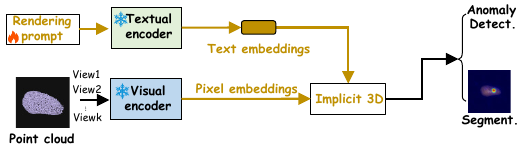}%
    \label{fig2:PointAD}}
    \\
    \subfigure[The schematic of PointAD+.]{\includegraphics[width=1\columnwidth]{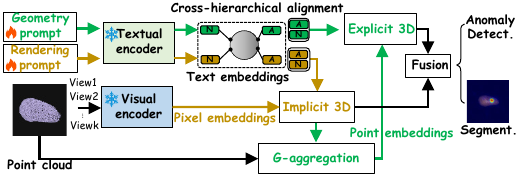}    \label{fig2:PointAD+}}
 \caption{Schematic for PointAD and PointAD+.}
 \label{f2: motivation}
\end{figure}

This paper proposes a unified framework to transfer the 2D knowledge of CLIP to 3D understanding for ZS 3D anomaly detection. We first present PointAD (conference version), which understands 3D anomalies by capturing rendering abnormality: \textbf{(1)} deriving 2D representations of point clouds via CLIP by rendering them from multiple views, \textbf{(2)} obtaining 3D prediction by aggregating the 2D predictions according to the correspondence between points and pixels. Specifically, as illustrated in Figure~\ref{fig2:PointAD}, PointAD models 3D abnormality by leveraging point-pixel relationships. Since pixel anomalies are 2D projections from 3D point anomalies, we refer to them as implicit 3D abnormality. These pixel anomalies contribute to point anomaly modeling from two perspectives: (1) pixels in different views reflect partial information of the same point; (2) pixels corresponding to the same point should exhibit similar semantics. After this, we introduce hybrid representation learning to optimize the rendering prompts from both 2D and 3D for implicit point representation. Specifically, to fully capture view-specific representations for 3D anomaly understanding, we treat each view as a separate 2D anomaly detection task. These 2D tasks across multiple views jointly facilitate the point anomaly learning from a multi-task learning (MTL) perspective. On the other hand, since pixel representations from the same point can be regarded as different instances, we aggregate their prediction results via multiple instance learning (MIL) to represent 3D anomalies.

\textcolor{orange}{
PointAD relies on view-wise 2D renderings to represent points. When point anomalies are not clearly reflected in the rendered views, the implicit point representations struggle to capture them. Such anomalies require to be capable of underlying spatial relations, yet CLIP is inherently limited in modeling the geometric structures present in point clouds. To overcome this limitation, we introduce explicit point representations that directly encode geometric information. Further, we propose PointAD+ that learns hierarchical representations: (1) align implicit point representations on the rendering layer, which interpret each point through aggregated pixel-level logits from multi-view renderings; and (2) align explicit point representations on the geometry layer, aggregated from implicit point representations and explicitly encoding spatial geometry.} 

 In Figure~\ref{fig2:PointAD+}, On the rendering layer, we use hybrid representation learning, as utilized in PointAD, for implicit point learning. On the geometry layer, we propose explicit point representation learning that reveals abnormal spatial relations by incorporating geometric information. To achieve this, we explicitly aggregate implicit point representations and propose G-aggregation to incorporate geometric information by representing point features via their neighboring features in 3D space. This allows the aggregated 3D representations to become spatially aware and effectively model abnormal spatial relationships. In addition, we consider the anomaly semantics on the rendering and geometry layers to stem from the same root cause. As a result, these layers can complement and learn from one another. Hence, we establish a cross-hierarchy contrastive alignment that brings the anomaly semantics from both layers together while pushing normal and anomalous semantics apart. This facilitates mutual anomaly learning across both hierarchies. During the test, we fuse the implicit and explicit 3D anomaly semantics for comprehensive anomaly modeling.

Benefiting from the alignment between point and pixel anomaly semantics, PointAD and PointAD+ can seamlessly incorporate additional RGB information and perform ZS multimodal 3D (multimodal 3D) detection without the need for extra modules or retraining. Extensive experiments across datasets with 32 class semantics demonstrate the superiority of PointAD and PointAD+ on ZS 3D and multimodal 3D anomaly detection and segmentation, in both one-class and cross-dataset settings. PointAD+ shows the performance improvement compared to PointAD. We also provide a comprehensive analysis of key modules' effectiveness and hyperparameters' robustness, including rendering quality and lighting conditions. The main contributions of this paper are summarized as follows:
\begin{itemize}[itemsep=2pt,topsep=0pt,parsep=0pt]
    \item  To the best of our knowledge, we take the first attempt to simultaneously investigate the challenging yet valuable ZS 3D and multimodal 3D anomaly detection. We propose to leverage the strong generalization of CLIP to detect and segment 3D anomalies over diverse unseen objects.
    \item We characterize 3D anomalies in two aspects: implicit and explicit 3D abnormality. To achieve a holistic understanding, we propose hierarchical representations learning to renovate the flat prompt into hierarchical prompts to capture implicit and explicit 3D abnormality. A cross-hierarchy alignment module is further introduced to facilitate mutual anomaly learning across layers.
    \item This paper presents a novel unified framework for ZS 3D anomaly detection, introducing two variants: PointAD and PointAD+. PointAD focuses on capturing implicit 3D abnormality, and PointAD+ expands on this by incorporating both implicit and explicit 3D abnormality for comprehensive anomaly learning.     \item Benefiting from the alignment of point
    and pixel anomaly semantics, PointAD and PointAD+ not only detect and segment 3D anomalies but also seamlessly integrate 2D RGB information in a plug-and-play manner during testing.
    \item Comprehensive experiments show that PointAD+ and PointAD outperform the baselines in ZS 3D and multimodal 3D by a large margin. We hope that our models will serve as a springboard for future research on ZS 3D and multimodal 3D anomaly detection. 
\end{itemize}

\textcolor{orange}{
\textbf{Discussion:} Compared to the conference version, the main extended contributions of the journal conference are as follows:\textbf{ Field contribution:} PointAD+ is the first to categorize point learning in ZS 3D anomaly detection into implicit point learning and explicit point learning. It highlights that CLIP-based frameworks fail to capture the geometric information of point clouds, which is crucial for accurate 3D anomaly detection; \textbf{Technical contribution:} PointAD+ proposes an effective approach to explicitly model point-wise representations from 2D features and incorporate geometric information into point representations. Then, PointAD+ introduces hierarchical representation learning to simultaneously align implicit and explicit point representations. PointAD+ achieves consistent improvements across all datasets in the paper; \textbf{Research contribution:} This work opens a new avenue for future research by encouraging the integration of stronger point-learning methods, such as PointNet-style or DGCNN-style architectures, to more effectively embed spatial relations into 2D representations for structural 3D anomalies.
}

\section{Related Work}
\subsection{3D Anomaly Detection}
3D anomaly detection has recently gained attention in the community of anomaly detection. Unlike 2D anomaly detection, which leverages RGB information, 3D anomaly detection focuses on identifying unusual spatial relationships in 3D objects. While 2D anomaly detection has been extensively explored, research on 3D anomaly detection remains relatively limited. This disparity is largely due to the challenges associated with acquiring high-quality 3D data, as collecting 3D datasets is more costly and labor-intensive compared to 2D images.
MVTec 3D-AD~\cite{bergmann2023anomaly} is the first public dataset, which provides the point cloud anomalies and the corresponding 2D-RGB information. Subsequently, Eyecandies~\cite{Bonfiglioli_2022_ACCV} and Real3D-AD~\cite{liu2023realdad} provide the 3D dataset for a wide exploration~\cite{li2024towards}. These datasets bridge the connection between 3D and 2D anomaly detection. 3D-ST~\cite{bergmann2023anomaly} uses a teacher net to extract dense local geometry descriptors and design a student net to match such descriptors. AST~\cite{rudolph2023asymmetric} introduces asymmetric teacher and student net to further improve 3D anomaly detection. Instead of only using point clouds, BTF~\cite{horwitz2023back}, M3DM~\cite{wang2023multimoda, wang2025m3dm},  CPFM~\cite{cao2023complementary}, and SDM~\cite{chu2023shape} integrate point features and RGB pixel features to detect 3D anomalies. While these approaches exhibit commendable performance by storing object-specific normal point and pixel features within the unsupervised learning framework, such paradigms simultaneously limit their generalization capacity to point clouds from unseen objects, which is crucial to detecting anomalies when the target object is unavailable. To fill this gap, we introduce PointAD+ and PointAD to identify unseen anomalies across diverse objects. They extend CLIP to the realm of ZS 3D and show robust generalization in capturing generic normality and abnormality within points. Furthermore, PointAD+ and PointAD serve as a unified framework, allowing seamless integration of 3D and RGB without additional training.

\vspace{-0.5em}
\subsection{3D Feature Extraction} 
The approaches for representing 3D objects can be broadly categorized into two branches. One way directly employs point-based networks to model 3D objects. Methods like PointNet~\cite{qi2017pointnet} and PointNet++~\cite{qi2017pointnet++} model the relationships between individual points to extract 3D features. The other way involves converting 3D data into a 2D format~\cite{su2015multi, goyal2021revisiting} by rendering point clouds into multiple views. This allows the use of robust 2D backbones to process the 2D renderings and subsequently handle 3D information. PointCLIP~\cite{zhang2022pointclip} directly projects raw points onto image planes for efficiency, however, this method often results in depth maps that lack detailed geometry information. Conversely, rendering-based methods~\cite{su2015multi, hamdi2021mvtn} generate high-quality 2D renderings from point clouds, preserving detailed local semantics. More related work in unsupervised 3D anomaly detection includes 3DSR~\cite{zavrtanik2024cheating}, which extracts depth images as 3D information, and CPFM~\cite{cao2023complementary}, which utilizes normal features from 2D renderings. In this paper, we adopt a rendering strategy for source point clouds and their annotations to obtain high-quality 2D rendering information and corresponding supervision. This enables PointAD+ and PointAD to capture generic anomaly semantics better, facilitating the recognition of abnormality in unseen objects.

\vspace{-0.5em}
\subsection{Prompt Learning} With the increase in model parameters, training a model from scratch requires substantial computational resources and time. This high training cost hinders the rapid development of deep learning technologies. Therefore, rather than training or fine-tuning the entire network, prompt learning focuses on optimizing a minimal set of parameters to adapt the network for downstream tasks. CoOp, a seminal work in prompt learning~\cite{zhou2022learning, zhou2022conditional}, introduces global context optimization to update learnable text prompts for few-shot recognition. DenseCLIP~\cite{rao2022denseclip} extends this approach to dense classifications. More recently, AnomalyCLIP~\cite{zhou2023anomalyclip} introduces object-agnostic prompt learning to capture generic normality and abnormality in images for 2D anomaly detection. Other works, such as~\cite{cao2024adaclip} and~\cite{qu2024vcp}, have built upon this by incorporating additional vision information. In contrast, PointAD+ innovatively transforms the flat prompts used in these methods into hierarchical prompts to capture both rendering and spatial abnormality for ZS 3D anomaly detection. Leveraging this hierarchical structure, we introduce hierarchy representation learning to achieve an in-depth understanding of ZS 3D and multimodal 3D anomaly detection.

\section{PointAD+}

\begin{figure*}[t]
  \centering
    \subfigure[Explicit point learning is superior to Implicit point learning
]{\includegraphics[width=0.33\textwidth]{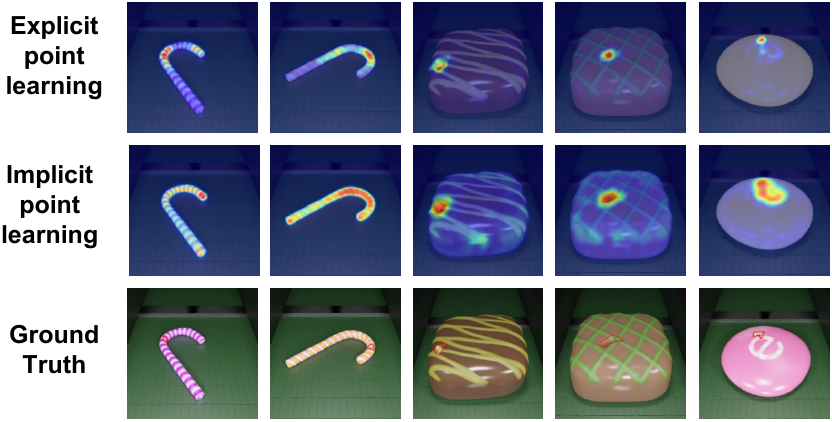}%
          \label{fig: ei1}}
    \hfil
    \subfigure[Implicit point learning is superior to explicit point learning
]{\includegraphics[width=0.33\textwidth]{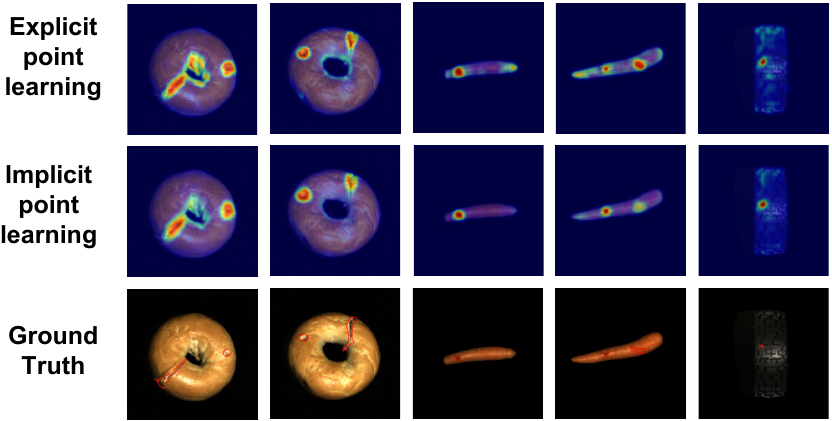}%
    \label{fig: ei2}}
 \hfil
    \subfigure[Performance improvement of PointAD+ over PointAD on stereo point clouds (Real3D-AD)
]{\includegraphics[width=0.33\textwidth]{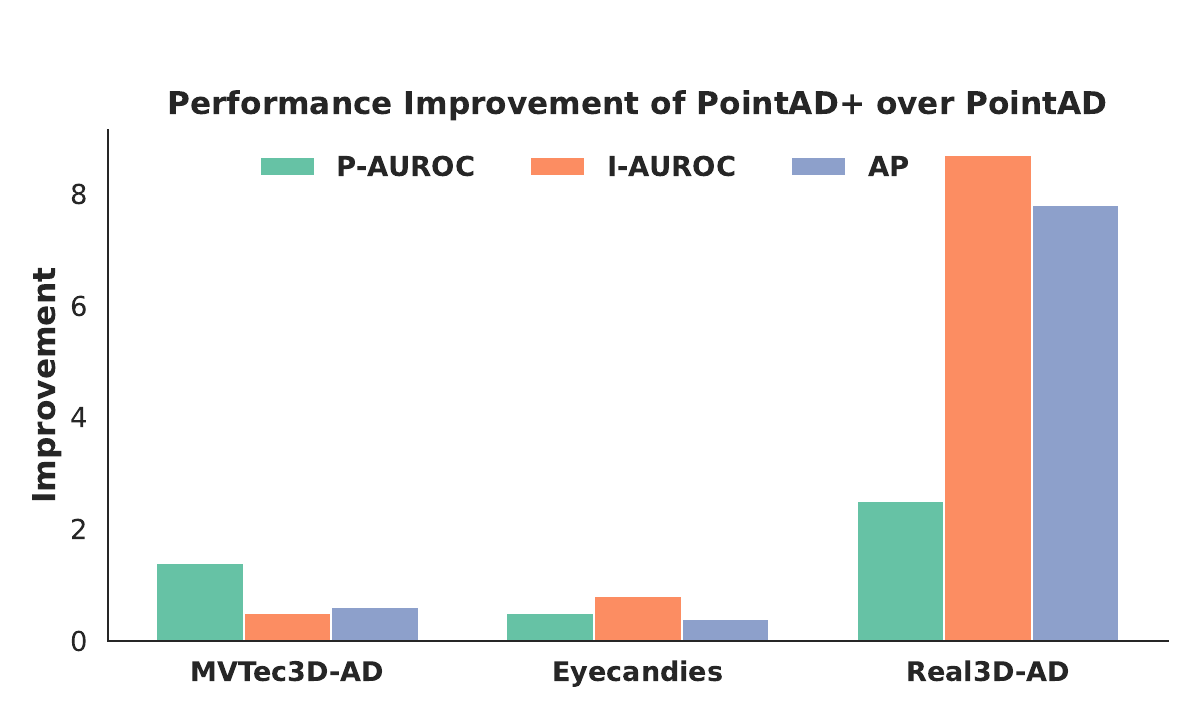}%
    \label{fig: ei3}}
\label{fig: ei12}
\caption{ \textcolor{orange}{Visualization of explicit and implicit point score maps. \textbf{Top row:} point segmentation using explicit point learning;  \textbf{Middle row:} point segmentation using implicit point learning; \textbf{Bottom row:} ground truths.}}
\end{figure*}

\subsection{A Review of CLIP and AnomalyCLIP}
CLIP, a representative VLM, aligns visual representations with the corresponding textual representations, where an image is classified by comparing the cosine similarity between its visual representation and textual representations of given class-specific text prompts. 
Specifically, given an image $x_i$ and target class set $\mathcal{C}$, visual encoders output the global visual representation $f_i \in \mathbb{R}^d$ and local visual representations $f_i^m \in \mathbb{R}^{h \times w \times d}$, where $h$, $w$, and $d$ are the height, width, and dimension, respectively. Textual representations $g_c$ are encoded by textual encoder $\mathcal{T}$ with the commonly used text prompt template \texttt{A photo of a [c]}, where $ c \in \mathcal{C}$. The probability that $x_i$ belongs to $c$ can be computed as:
\begin{equation}
     P(g_c, f_i) = {\frac{exp(cos(g_c, f_i)/\tau)}{\sum_{c\in \mathcal{C}}exp(cos(g_c, f_i))/\tau)}},
\label{equ: softmax}
\end{equation}
where $cos(\cdot, \cdot)$ and $\tau$ represent the cosine similarity and temperature used in CLIP, respectively. The segmentation $S_{i(c)} \in \mathbb{R}^{h\times w}$ for class $c$ can be computed as $Seg(g_c, f_i^m)$, where each entry ($m$,$v$) is calculated as $P(g_c, f_{i,u,v}^m)$. For clarification, $g_a$ and $g_n$ represent the text embeddings for the abnormality class and the normality class, respectively. When a class is not specified, we use $g_c$ to refer to the text embeddings containing all classes.

\textcolor{orange}{
Replacing fixed text prompts with learnable text prompts is an effective method to adapt CLIP for downstream tasks. AnomalyCLIP~\cite{zhou2023anomalyclip} introduces object-agnostic text prompts to adapt CLIP for ZS 2D anomaly detection. The success of AnomalyCLIP demonstrates the promise of adapting CLIP for ZSAD by utilizing a lightweight, learnable text prompt. However, ZS 3D anomaly detection remains underexplored, primarily because there is no 3D foundation model with the strong generalization capabilities of CLIP. In this paper, we explore how to transfer CLIP’s strong generalization ability to 3D anomaly detection in a ZS manner.}  

\subsection{Overview of PointAD and PointAD+}
\begin{figure*}[h]
\begin{center}
\centerline{\includegraphics[width=1\textwidth]{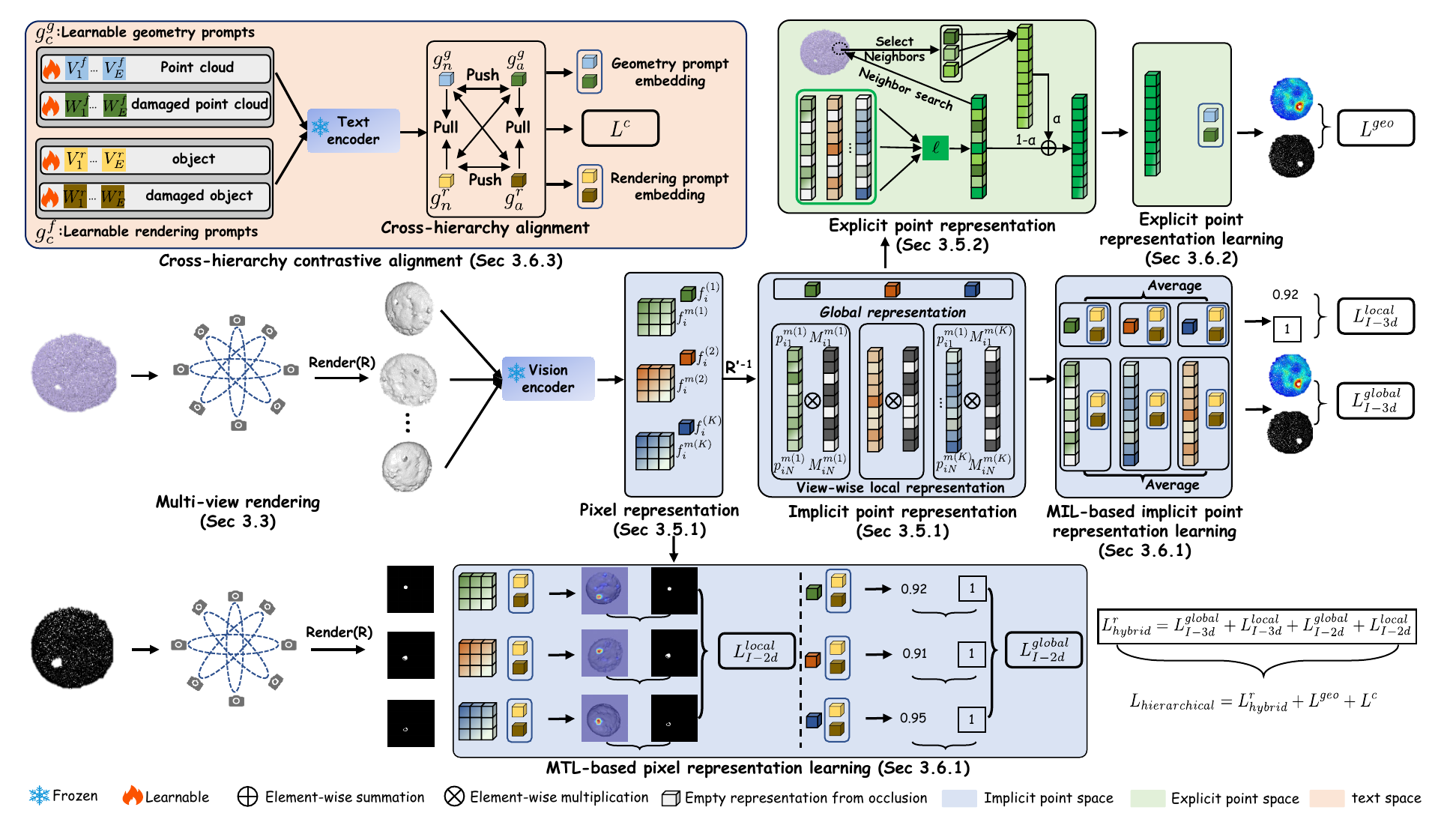}}
\caption{Framework of PointAD+. PointAD+ interprets 3D abnormality through implicit and explicit 3D abnormality. For implicit 3D abnormality, CLIP's vision encoder extracts 2D global and local representations from the renderings, and then the resulting 2D representations are projected into implicit point representation to capture point anomaly semantics (rendering layer). For explicit 3D abnormality, we propose G-aggregation to obtain the explicit point representation by aggregating the implicit point representation and then incorporating geometric information (geometry layer). Holding these layers, cross-hierarchy alignment is further introduced to facilitate mutual learning across layers. Finally, hierarchical representation learning jointly optimizes the text embeddings to align explicit and implicit point representations with learnable rendering and geometric prompts, capturing generic anomaly patterns comprehensively.}
\label{f2:overview}
\end{center}

\end{figure*}

ZS AD requires models to identify anomalies in previously unseen 3D objects with diverse class semantics. Our goal is to transfer the strong 2D generalization ability of CLIP from images to 3D point cloud recognition.
\textcolor{orange}{
As illustrated in Figure~\ref{f2:overview}, PointAD+ first interprets point clouds through (1) implicit point representations, which capture 3D abnormality via pixel features from 2D renderings (Section~\ref{Implicit point space}), and (2) explicit point representations, which aggregate implicit point representations for point-level representation and encode the geometry information via G-aggregation. (Section~\ref{Explicit point space})
}

We then propose hierarchical representation learning (Section~\ref{Hierarchical representation learning}), which aligns these representations with learnable hierarchical text prompts, enabling the model to incorporate generic anomaly semantics from rendering-based abnormality (rendering (bottom) layer) and geometry-based abnormality (geometry  (top) layer). On the rendering layer (Section~\ref{Implicit point representation learning on the rendering layer (PointAD)}), we introduce hybrid representation learning to align rendering prompts with both pixel representations from 2D renderings and implicit point representations in the implicit point space. On the geometry layer (Section~\ref{Explicit point representation learning on the geometry layer}), we align geometric prompts with explicit point representations, which are derived from implicit representations and further are encoded in geometry information, in explicit point space.

Built on these two layers, we further enhance the overall understanding of PointAD+ through cross-hierarchy alignment, which promotes mutual anomaly learning across rendering and geometry layers (Section~\ref{Cross-hierarchy contrastive alignment}). Finally, we integrate anomaly semantics from both hierarchies to generate the final outputs. Moreover, thanks to explicit alignment between 3D and 2D representations, both PointAD+ and PointAD can directly incorporate RGB information during testing, enabling ZS multi-modal 3D anomaly detection.

\textcolor{orange}{
\subsection{Motivation of PointAD+}
\label{sec:Motivation of PointAD+}
PointAD employs implicit point representations to capture 3D anomaly semantics by aggregating pixel-level abnormalities into point-level ones. However, when anomalies are not well reflected in renderings, this approach can lead to missed anomalies and false alarms (Figure~\ref{fig: ei1}). In contrast, explicit point segmentation encodes geometric information to localize anomaly regions with higher precision, as it is more sensitive to spatial variations. Yet, explicit point learning is not universally superior: as shown in Figure~\ref{fig: ei2}, it struggles with edge points, where abrupt spatial changes often trigger false alarms. Implicit point representations, by relying on 2D features, are less sensitive to such edge variations and avoid this problem. To exploit the strengths of both, PointAD+ introduces hierarchical representation learning that integrates implicit point learning on the rendering layer and explicit point learning on the geometry layer, enabling a more comprehensive understanding of abnormalities. Figure~\ref{fig: ei3} shows that PointAD+ achieves a larger performance gain on Real3D-AD (stereo point clouds) than on MVTec3D-AD and Eyecandies. Real3D-AD contains stereo point clouds, which require the model to be more sensitive to point relations. The significant improvement demonstrates the superiority of PointAD+ in spatial relation modeling.
}

\subsection{Multi-view Rendering}
\vspace{-0.2em}
Multi-view projection is a crucial technique for understanding point clouds from 2D perspectives. Some multi-view projection approaches project point clouds into various depth maps, providing adequate shape information for class recognition~\cite{zhang2022pointclip}. However, in this paper, the objective of PointAD+ and PointAD is to learn both generic global and local anomaly semantics. Depth-map projection lacks sufficient resolution to represent fine-grained anomaly semantics accurately. Hence, we adopt high-precision rendering to preserve the original 3D information offline. Specifically, given an auxiliary dataset of point clouds $\mathcal{D}_{3d} = \{(x^{3d}_i, y^{3d}_i)\}_{i=1}^N$, we define the rendering matrix as $R^{(k)}$ for the $k$-$th$ view, with a total of $K$ views. We simultaneously render point clouds and point-level ground truths from different views to obtain their corresponding 2D renderings, which is given by $x_{i}^{(k)} = R^{(k)}(x^{3d}_i)$ and $y_{i}^{(k)} = R^{(k)}(y^{3d}_i)$, where $x_{i}^{(k)} \in \mathbb{R}^{H\times W}$ and $y_{i}^{(k)} \in \mathbb{R}^{H\times W}$ respectively represent the $k$-$th$ 2D rendering and corresponding pixel-level ground truth in the $i$-$th$ point cloud, i.e., 0 for normality and 1 for abnormality.
 
\vspace{-0.5em}
\subsection{Pixel and Point representation}
\textcolor{orange}{
PointAD+ interprets the 3D abnormality in implicit point space and explicit point space. 
In \textbf{implicit point space}, PointAD+ models pixel representation extracted by CLIP and implicit point representation that implicitly models point abnormality through point-pixel relationships. In \textbf{explicit point space}, PointAD+ explicitly models the point-level representation by aggregating implicit point representation and using G-aggregation to incorporate geometric information.}
\subsubsection{Implicit point space}
\label{Implicit point space}
Rendering abnormality in implicit point space reflects on pixel representation and implicit point representation.

\textbf{Pixel representation:} For a point cloud $x^{3d}_i$, we first obtain the 2D renderings $\mathcal{X}_i = \{x_{i}^{(k)}\}_{k=1}^K$. Then, these renderings are encoded via the vision encoder of CLIP to obtain \textbf{pixel representation:} global pixel representations \(\mathcal{F}_i \hspace{-0.2em}=\hspace{-0.2em} \{f_{i}^{(k)}\}_{k=1}^K\), and local pixel representations $\mathcal{F}_i^{m} \hspace{-0.2em}=\hspace{-0.2em} \{f_i^{m(k)}\}_{k=1}^{K}$. 

\textbf{Implicit point representation:} The consistency constraint requires points and the corresponding pixels to maintain the same semantics. Therefore, the global implicit point representation $p$ and local implicit point representations ${p}^{m}$ are expected to include their corresponding pixel representations in each view. Formally, let us define the global implicit point representation in the $i$-$th$ point cloud as $p_{i} = \{p_{i}^{(k)} | p_{i}^{(k)} = f_{i}^{(k)}\}_{k = 1}^{K}$. Also, we denote the local implicit point representation:
\begin{equation}
p^m_{i} = \{p^{m(k)}_{i} | p^{m(k)}_{i} = \{p^{m(k)}_{i,1}, p^{m(k)}_{i,2}, \cdots, p^{m(k)}_{i,n}\}\}_{k=1}^K, 
\end{equation}
where $p^{m(k)}_{i,j} = f_{i,u,v}^{m(k)}$ with $(u, v) = R'^{(k)}(a_{i,j}, b_{i,j}, c_{i,j})\}$. ($a_{i,j}$, $b_{i,j}$, $c_{i,j}$) represents the 3D coordinate of the $j$-$th$ point of $i$-$th$ point cloud. $R'^{(k)}$ is the transformation between the point and pixel space, derived as $R'^{(k)} = \frac{h}{H}R^{(k)}$. 

Points at different positions may correspond to different numbers of pixel representations as other points hide them from a specific viewpoint. In this case, we introduce a view-wise visibility mask $M$, where $M_{i,j}^{k}$ indicates whether the $j$-$th$ point of the $i$-$th$ point cloud is visible in the $k$-$th$ view. We compare the point depth projected into the same pixel in the same view and set the corresponding visibility mask to 1 for the point with the minimum depth, and to 0 for the other points. Let $\mathcal{Q}_{i,u,v}^{(k)}$ denote the depth set of all points that are projected into the same pixel indexed by $(u,v)$ in the $i$-$th$ point cloud in the $k$-$th$ view. $\mathcal{Q}_{i,u,v}^{(k)}$ and $M_{i,j}^{k}$ are respectively given as 
$\mathcal{Q}_{i,u,v}^{(k)} = \{c_{i,j} \mid R'^{(k)}(a_{i,j}, b_{i,j}, c_{i,j}) = (u, v)\}_{j=1}^{n}$ and $M^{(k)}_{i,j} = \mathbb{I}(i,j,k=\argmin \limits_{i,j,k}\{c_{i,j} \mid c_{i,j} \in \mathcal{Q}_{i,u,v}^{(k)}\})$,
where $\mathbb{I}(\cdot)$ is an indicator function. Thus, $p^{m(k)}_{i}$ is reformulated as: $p_{i}^{m(k)} = \{p^{m(k)}_{i,j}*M^{(k)}_{i,j}\}_{j=1}^{n}$.
\subsubsection{Explicit point space}
\label{Explicit point space}
Since we interpret 3D abnormalities using 2D abnormalities through a 2D encoder in the implicit point space, this approach fails to consider the intrinsic geometry of the point cloud. 
Therefore, we seek to model explicit point representations to better capture geometric information within point clouds.

\textbf{Explicit point representation:}
Explicit point representation requires obtaining the specific point representation for modeling abnormal spatial positions, rather than set-like implicit point representation. Since point information is represented as its counterpart pixels in our framework, we resort to aggregating the above implicit point representation to obtain explicit point representation:
\begin{equation}
q_{i}^m = \{q_{ij}^m | q_{ij}^m = \ell(p_{ij}^{m(1)}, ..., p_{ij}^{m(K)})\}_{j = 1}^n, 
\end{equation}
where $\ell(\cdot)$ represents the aggregation strategy, i.e., take the mean or maximum of the pixel representations as point representations or employing a parameterized network. If a parameterized network is adopted, it would bring additional computational overhead. To ensure computational efficiency, we adopt the mean operation as the aggregation strategy. 

\textbf{G-aggregation} The above representations lack the inherent geometry information and further reduce the gap between the 2D and 3D representations. This makes it challenging for the point representation to reflect the relative spatial relationships within point clouds comprehensively. Hence, we propose G-aggregation to inject the geometry information into point representation aggregation, enabling the point representations to be aware of geometry characteristics. In G-aggregation, the process begins with an initial aggregation, where the explicit point representation is obtained by aggregating the implicit point representation. To further refine the initial explicit point representation with geometry details, we search the k nearest neighbors of $q_{ij}^m$ within distance space $\mathcal{G}$, i.e., $
\mathcal{N}_{x_{ij}^{3d}} = \{x_{il}^{3d} \mid x_{il}^{3d} \in \operatorname{TopK}(\{x_i^{3d}\} \setminus x_{ij}^{3d}, \mathcal{G}(x_{ij}^{3d}, x_{il}^{3d}))\}$. In this work, we adopt Euclidean distance as the primary metric for measuring relative distances. When the neighboring points are selected, the second aggregation step is performed to refine the point representations. This is achieved by weighting the point representations of their neighbors, using a Gaussian kernel $\mathcal{K}(x_{ij}^{3d}, x_{il}^{3d}) = \exp\left(-\|x_{ij}^{3d} - x_{il}^{3d}\|^2/2\sigma^2\right)$ to assess their contributions, followed by normalization.

\begin{footnotesize}
\textcolor{orange}{
\begin{align}
\tilde{\mathcal{K}}(x_{ij}^{3d}, x_{il}^{3d}) = \frac{\mathcal{K}(x_{ij}^{3d}, x_{il}^{3d})}{\sum_{x_{in}^{3d} \in \mathcal{N}_{x_{ij}^{3d}}} \mathcal{K}(x_{ij}^{3d}, x_{in}^{3d})},& \enspace
\tilde{q_{ij}^m} \hspace{-0.05cm}= \hspace{-0.3cm}\sum_{q_{il}^m \in \mathcal{N}_{q_{ij}^m}} \hspace{-0.25cm}\tilde{\mathcal{K}}(x_{ij}^{3d}, x_{il}^{3d}) \cdot q_{il}^m, 
\end{align}
}
\end{footnotesize}
In doing so, we represent the point using its neighboring features, allowing it to incorporate geometric information. As a result, the derived representations exhibit spatial representation consistency. Finally, the final point representation is obtained by fusing the point representations from two aggregation stages, weighted by a scalar parameter $\alpha$. G-aggregation enables point representations to balance the specificity of individual points and their spatial continuity for more accurate segmentation.

\begin{footnotesize}
\begin{align}
\textcolor{orange}{
\hat{q_{ij}^m} = \alpha \tilde{q_{ij}^m} + (1-\alpha)q_{ij}^m. 
}
\end{align}
\end{footnotesize}

\subsection{Hierarchical representation learning}
\label{Hierarchical representation learning}
The key to ZS 3D anomaly detection lies in the model's ability to capture generic anomaly semantics, enabling it to detect anomalies across various objects. The above section focuses on effectively representing 3D anomalies. 
\textcolor{orange}{
Here, we propose hierarchical representation learning to align these representations on two layers: the rendering (bottom) layer aligns the implicit point representation and pixel representation with the learnable rendering prompt in implicit point space; the geometric (top) layer aligns the explicit point representations with the learnable geometric prompts in explicit point space. Furthermore, we introduce cross-hierarchy alignment to facilitate mutual anomaly learning across layers. This enables PointAD+ to produce generic text embeddings for generalized anomaly semantics. 
}

Specifically, we design two types of learnable text prompts: rendering prompts $g^r_c$ and geometry prompts $g^g_c$, each comprising both normality and abnormality components. Rendering prompts are designed to learn implicit point anomalies by leveraging point-pixel correspondence to represent points through their associated rendering pixel representations. In contrast, geometry prompts capture abnormal spatial relationships by aligning aggregated point representations enriched with geometric information within point clouds. PointAD initializes these prompts in AnomalyCLIP~\cite{zhou2023anomalyclip} manner:

\vspace{-0.6em}
\begin{footnotesize}
\begin{equation}
\begin{rcases}
\begin{rcases}
    g^r_c = \left\{
        \begin{aligned}
    &t^r_n =[V^r_1]\dots[V^r_E][object], \\&t^r_a =[W^r_1]\dots[W^r_E][damaged][object]    \end{aligned} \right. \nonumber 
\end{rcases} \text{PointAD}\\
    \hspace{-5cm}g^g_c = \left\{
        \begin{aligned}
    &t^g_n =[V^g_1]\dots[V^g_E][point] [cloud], \\&t^g_a =[W^g_1]\dots[W^g_E][damaged][point][cloud]  \hspace{+0.5cm}  
    \end{aligned} \right. \nonumber
\end{rcases} 
\text{PointAD+}
\end{equation}
\end{footnotesize}%
where $V$ and $W$ are learnable word embeddings, respectively. PointAD+-CoOp instead initializes the prompts by replacing \texttt{object} and \texttt{point cloud} with the specific class name.

\subsubsection{\textbf{Implicit point representation learning on the rendering layer (PointAD)}}
\label{Implicit point representation learning on the rendering layer (PointAD)}
Rendering anomalies in 2D renderings implicitly represent 3D anomalies. Since a single point could be projected into different views, the pixel representations corresponding to the same point should share similar semantics and collectively characterize the point. Additionally, learning to identify 2D-pixel anomalies in each rendering can contribute to more effective 3D anomaly recognition. Combining these two aspects, we propose hybrid representation learning to optimize the rendering prompts for implicit point representation. This forms PointAD.

\textbf{MIL-based implicit point representation learning}. To fully incorporate rendering anomaly semantics into PointAD+, we respectively devised two losses to regularize the rendering anomalies from 3D global anomaly and local anomaly semantics. First, we compute the cosine similarity between rendering textual representations $g^r_c$ and global pixel representations $f_i$ in each view. As point clouds are projected from different views, the resulting renderings in each view reflect specific parts of point clouds. We use view-wise MIL to integrate global implicit representations $p$ and then align global labels to capture the global semantics. Formally, the global 3D loss is defined as:
\textcolor{orange}{
\begin{equation}
    \textstyle L_{I(3d)}^{global} = \frac{1}{N}  \sum_{i}\textrm{CE}(\frac{1}{K} \textstyle \sum_{f_{i}^{(k)}\in p_i} P(g^r_c, f_{i}^{(k)}), \max{(y^{3d}_i)}).
\end{equation}
}
where $\textrm{CE}(\cdot)$ represents the cross entropy. As for local anomaly semantics, we quantify the cosine similarity between textual representations $g^r_c$ and local implicit point representations $p^m$. Considering that one point would be projected into pixels from different views, each pixel representation belonging to the same point is regarded as multiple instances. We adopt the pixel-wise MIL to achieve the consistent constraint. The implicit point segmentation can be formulated mathematically as follows:
\textcolor{orange}{
\begin{align}
    \textstyle  S_{i(a)}^{3d} &= \frac{1}{K} \sum_{k}Seg(g^r_a, p_i^{m(k)})\\ 
    \textstyle  S_{i(n)}^{3d} &= \frac{1}{K}\sum_{k}Seg(g^r_n,  p_i^{m(k)}).
\end{align}
}
However, deriving such implicit point segmentation requires similarity computation for each point. It brings a significant memory burden, with a huge computational complexity of \( O(Knd) \), which is unaffordable for a single NVIDIA RTX 3090 24GB GPU. To address this computational challenge, we resort to the rendering correspondence between points (3D space) and their corresponding pixels within each view (2D space). Take the anomaly segmentation for example, we first can rewrite implicit point segmentation from the view perspective as $S_{i(a)}^{3d} =\frac{1}{K} \textstyle \sum_{k}S_{i(a)}^{3d(k)}$. Then, 
the $k$-$th$ division of implicit point segmentation can be transformed into the 2D counterpart through the rendering projection $S_{i(a)}^{3d(k)}=(R^{(k)})^{(-1)}S_{i(a)}^{2d(k)} \otimes M_i^{(k)}$, where $\otimes$ is the Hamiltonian product. The $k$-$th$ 2D counterpart can be computed as:
\begin{equation}
S_{i(a)}^{2d(k)} = \textrm{Up}(Seg(g_a, f_{i}^{m(k)})),
\label{equ:rendering2d}
\end{equation}
where the operator $\textrm{Up}(\cdot)$ represents bilinear interpolation from feature space to 2D space. Finally, we can reformulate the implicit point segmentation as follows: 
\begin{align}
\hspace{-0.3cm}\textstyle S_{i(a)}^{3d} \hspace{-0.05em}= \hspace{-0.05em}\frac{1}{K} \sum_{k}\Big ((R^{(k)})^{(-1)}\textrm{Up}(Seg(g^r_a, f_{i}^{m(k)})) \otimes M_i^{(k)}\Big). 
\label{equ:rendering3d}
\end{align}
From the equation, we can observe that primary computation can be conducted in feature space, with a computational complexity of \( O(Khwd) \). This is a substantial overhead reduction compared to \( O(Knd) \) since feature space is much smaller than 3D space, \ie, $ h\times w \ll n $, where $h\times w = 24 \times 24 = 576$ but $n = 336\times336 = 112896$. After that, Dice Loss is employed to model the decision boundary of anomaly regions precisely. Let $I$ represent a full-one matrix of the same size as $y_{i}^{3d}$. Implicit 3D local loss $L_{3d}^{local}$ is defined as:
\textcolor{orange}{
\begin{equation}
\hspace{-0.1cm} \textstyle L_{I(3d)}^{local}=\frac{1}{N} \sum_{i}\Big(\textrm{Dice}(S_{i(n)}^{3d}, I - y_{i}^{3d}) + \textrm{Dice}(S_{i(a)}^{3d}, y_{i}^{3d})\Big). 
\end{equation}
}
\textbf{MTL-based pixel representation learning}.
We further improve PointAD+ and PointAD point understanding by capturing 2D glocal anomaly semantics into the object-agnostic text prompt template. We treat the anomaly recognition for one rendering from the point cloud as a task. Hence, we formulate the anomaly semantics learning for multiple 2D renderings as MTL. MTL-based 2D representation learning is divided into two parts for respective alignment to 2D global and local anomaly semantics.
For 2D global semantics, we use \textrm{CrossEntropy} to quantify the discrepancy between the textual representations and each global Pixel representation $f_i$. Global MTL-based 2D representation learning $L_{I(2d)}^{global}$ is defined as:
\textcolor{orange}{
\begin{gather}
    \textstyle L_{I(2d)}^{global} = \frac{1}{NK} \sum_{i, k}\textrm{CE}(P(g^r_c, f_{i}^{(k)}), \max{(y_{i}^{(k)})}). 
\end{gather}
}
Also, we focus on 2D abnormal regions to understand pixel-level anomalies. As the anomaly regions are typically smaller than normal regions, we employ \textrm{Focal} Loss to mitigate the class imbalance besides \textrm{Dice} Loss. Let $\oplus$ denote the concatenation operation. Local MTL-based 2D representation learning $L_{I(2d)}^{local}$ is given as follows: 
\begin{equation}
\begin{split}
\setlength{\abovedisplayskip}{2pt}
\textstyle L_{I(2d)}^{local} = \frac{1}{NK} \sum_{i,k} \textrm{Focal}(S_{i(n)}^{2d(k)} \oplus S_{i(a)}^{2d(k)}, y_{i}^{(k)}) + \\
\textrm{Dice}(S_{i(n)}^{2d(k)}, I - y_{i}^{(k)})+ \textrm{Dice}(S_{i(a)}^{2d(k)}, y_{i}^{(k)}). 
\end{split}
\end{equation}

In summary, rendering anomaly learning includes implicit point representation learning (3D) and MTL-based pixel representation learning (2D). We minimize $L^{r}_{hybrid}$ to incorporate global and local implicit 3D anomaly semantics:
\begin{align}
\textcolor{orange}{
    L^{r}_{hybrid}  = L_{I(3d)}^{global} + L_{I(3d)}^{local} + L_{I(2d)}^{global} + L_{I(2d)}^{local}.}
\end{align}

\subsubsection{\textbf{Explicit point representation learning on the geometry layer}}
\label{Explicit point representation learning on the geometry layer}
Although we learn implicit point representation from rendering pixel anomalies on the rendering layer, modeling the explicit point representation for spatial abnormality also has the same importance. We align explicit point representations with geometry prompts to capture point spatial relations. Specifically, the explicit point segmentation is computed as the similarity between the explicit point representation and geometry prompts. 
\vspace{-1em}
\begin{align}
\textcolor{orange}{
    \textstyle  T_{i(a)}^{3d} = Seg(g^g_a, \hat{q}^m_i), \label{equal:geometry3d}
    } \\
    \textcolor{orange}{
    \textstyle  T_{i(n)}^{3d} = Seg(g^g_n, \hat{q}^m_i).
    }
\end{align}
Here, to incorporate the abnormal spatial relations within point clouds, we minimize $L^{geo}$ to align explicit local point representation $q_i^m$ with the 3D anomaly semantics: 
\begin{equation}
\textcolor{orange}{
\hspace{-0.2cm} \textstyle L^{geo}=\frac{1}{N} \sum_{i}\Big(\textrm{Dice}(T_{i(n)}^{3d}, I - y_{i}^{3d}) + \textrm{Dice}(T_{i(a)}^{3d}, y_{i}^{3d})\Big). 
}
\end{equation}

\begin{figure}
  \centering
    \subfigure[Inference for ZS 3D anomaly detection]{\includegraphics[width=0.42\textwidth]{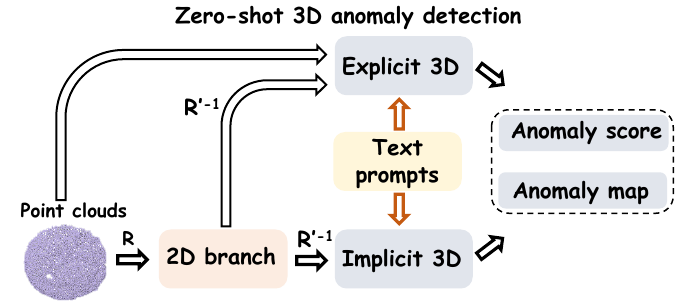}%
          \label{fig: inference_1}}
    \hfil
    \subfigure[Inference for ZS multimodal 3D anomaly detection]{\includegraphics[width=0.42\textwidth]{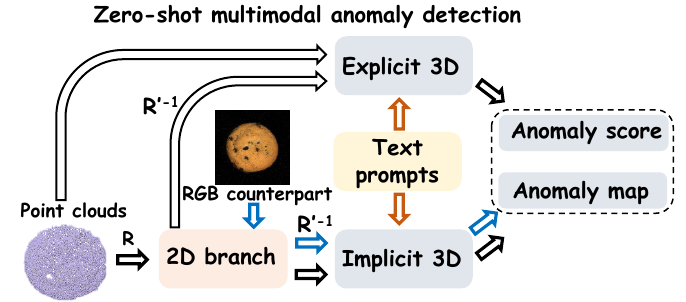}%
    \label{fig: inference_2}}
    \vspace{-1em}
 \caption{Inference schematic for ZS 3D and multimodal 3D detection.}
\end{figure}

\begin{figure*}[h]
\begin{center}
\centerline{\includegraphics[width=1\textwidth]{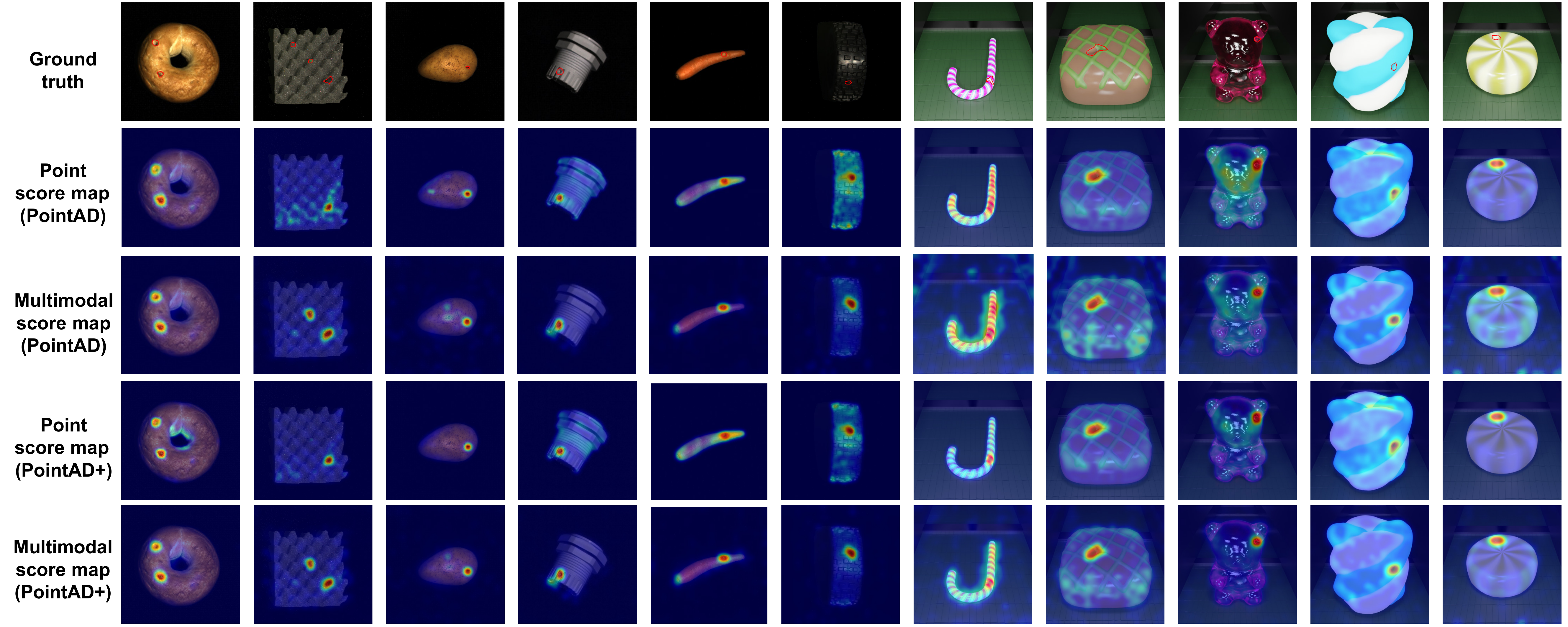}}
\caption{Visualization for PointAD and PointAD+ on MVTec3D-AD and Eyecandies in ZS 3D and multimodal 3D anomaly detection. Each row visualizes the anomaly score maps of objects from cross-category and cross-dataset scenarios using the model trained on "cookie" from MVTec3D-AD.}
\vspace{-2em}
\label{f5:visualization}
\end{center}
\end{figure*}

\subsubsection{\textbf{Cross-hierarchy contrastive alignment}}
\label{Cross-hierarchy contrastive alignment}
So far, the anomaly learning on the rendering layer and geometry layers is independent. Since Implicit 3D abnormality and explicit 3D abnormality have different emphases, they may have common and special patterns to complement each other. Based on this, we propose cross-hierarchy alignment to promote mutual anomaly learning between pixel and point anomalies, facilitating comprehensive anomaly understanding in PointAD+. We aim to achieve knowledge sharing among the same semantics and contrast them with different semantics. Specifically, we pull the anomaly text embeddings from two layers apart and push them far away from the normal text embeddings of both layers, resulting in symmetric contrastive loss: 

\vspace{-1em}
\begin{footnotesize}
\begin{align}
L^{c} = 
&-\log \frac{\exp(\cos(g^g_a, g^r_a))}
               {\exp(\cos(g^g_a, g^r_a))+\exp(\cos(g^g_a, g^r_n))}
\nonumber\\
&-\log \frac{\exp(\cos(g^g_n, g^r_n))}
               {\exp(\cos(g^g_n, g^r_a)) + \exp(\cos(g^g_n, g^r_n))}
\\
&-\log \frac{\exp(\cos(g^g_a, g^r_a))}
               {\exp(\cos(g^g_a, g^r_a))+\exp(\cos(g^g_n, g^r_a))}
\nonumber\\
&-\log \frac{\exp(\cos(g^g_n, g^r_n))}
               {\exp(\cos(g^g_a, g^r_n)) + \exp(\cos(g^g_n, g^r_n))}
\nonumber
\end{align}
\end{footnotesize}

\subsection{Training \& Inference}
\subsubsection{Training} PointAD+ incorporates 3D anomaly semantics into the hierarchical prompts thus integrating implicit point representation (rendering layer) and explicit point representation (geometry layer). During training, we only optimize tiny learnable prompts, and the original parameters of CLIP are frozen to maintain its strong generalization. Hierarchy representation learning consists of three parts, i.e., rendering loss, geometry loss, and cross-hierarchy alignment loss:
\begin{equation}
    L_{hierarchical} = L^{r}_{hybrid} + L^{geo} + L^{c}.
\end{equation}

\subsubsection{Inference} Since our model provides a unified framework to understand anomaly semantics from point and pixel, it can not only perform \textbf{ZS 3D anomaly detection} but also \textbf{multimodal 3D anomaly detection in a plug-and-pay way}. Next, we will introduce the inference process in detail: 

\textbf{ZS 3D inference.} As illustrated in Figure~\ref{fig: inference_1}, given a point cloud $x_{i}^{3d}$, PointAD+ sum the implicit point segmentation (in Eq.~\ref{equ:rendering3d}) and explicit point segmentation the 3D segmentation (in Eq.~\ref{equal:geometry3d}) as the anomaly/point score map: $A^m_i \hspace{-0.2em}=\hspace{-0.2em}G_{\sigma}(0.5S_{i(a)}^{3d} + 0.5T_{i(a)}^{3d}$), where $G_{\sigma}(\cdot)$ represents the Gaussian filter. The global anomaly score incorporates glocal anomaly semantics from spatial abnormality and rendering abnormality and is computed as $A_i^s = 0.5\Big(\frac{1}{K}\sum_{f_{i}^{(k)}\in \mathcal{F}_i }P(g_c, f_{i}^{(k)})\Big) + 0.5\max{(A^m_i)}$. Note that we integrate the anomaly map $A_i^{m}$ and $B_i^m$ as well as the anomaly score $A_i^{s}$ and $B_i^{s}$ to keep them within the range of $[0, 1]$. Compared to PointAD+, PointAD just considers the implicit 3D anomalies. Its anomaly score map is denoted as $B^m_i \hspace{-0.2em}=\hspace{-0.2em}G_{\sigma}(S_{i(a)}^{3d})$, and the anomaly score is computed as $B_i^s = 0.5(\frac{1}{K}\sum_{f_{i}^{(k)}\in \mathcal{F}_i }P(g_c, f_{i}^{(k)})+0.5\max{(B^m_i)})$.

\textbf{ZS multimodal 3D inference.} When the RGB counterpart is available for testing, PointAD+ could directly integrate RGB information by feeding RGB images to 2D branch to derive pixel representations. In Figure~\ref{fig: inference_2}, we project these pixel representations back to 3D branch to respectively compute the RGB anomaly score map and anomaly score as $ A^{m(rgb)}_i = P(g_c, f_{i}^{(rgb)})$ and $A_i^{s(rgb)} = G_{\sigma}(S_{i(a)}^{3d(rgb)})$. The final multimodal anomaly score map and anomaly score are defined as $A_i^{m(mod)} = G_{\sigma}\big(0.5A_i^m + 0.5A^{m(rgb)}_i\big)$ and $A_i^{s(mod)} = 0.5(0.5A_i^{s(rgb)}+0.5A_i^s)+ 0.5\max{(A_i^{m(mod)})}$. Also, the final multimodal anomaly score map and anomaly score for PointAD  are defined as $B_i^{m(mod)} = G_{\sigma}\big(0.5B_i^m + 0.5A^{m(rgb)}_i\big)$ and $B_i^{s(mod)} =0.5(0.5A_i^{s(rgb)}+0.5B_i^s)+0.5\max{(B_i^{m(mod)})}$.

\section{Experiment}
\subsection{Experiment Setup}
\textbf{Dataset.} We evaluate the performance of ZS 3D anomaly detection on three publicly available 3D anomaly detection datasets, MVTec3D-AD, Eyecandies, and Real3D-AD. MVTec3D-AD comprises 4147 point clouds across 10 categories. These objects exhibit diverse object semantics, including bagel, cable gland, carrot, cookie, dowel, foam, peach, potato, rope, and tire. The training dataset comprises 2656 normal point clouds, and the test dataset includes 948 normal and 249 anomaly point clouds, covering several anomaly types. Point-wise annotations are available for the point clouds. MVTec3D-AD also provides corresponding 2D-RGB image counterparts for the point clouds. We remove the background plane of point clouds in the whole dataset like~\cite{horwitz2023back}. Eyecandies also has 10 different classes and provides the corresponding 2D RGB information. The more recent Real3D-AD is a challenging dataset containing stereo point clouds with 12 categories, without the corresponding RGB information.  

\textbf{Evaluation setting and metric.} Since the training dataset of MVTec3D-AD contains only normal samples, we employ an object test dataset as the auxiliary dataset to fine-tune PointAD and evaluate ZS anomaly detection on the remaining objects. Specifically, we report the average results using different objects as auxiliary data: carrot, cookie, and dowel for MVTec3D-AD; confetto, LicoriceSandwich, and PeppermintCandy for Eyecandies; and seahorse, shell, and starfish for Real3D-AD. Additionally, we investigate more challenging cross-dataset generalization settings, where we use PointAD+ and PointAD trained from one dataset to test all objects in another dataset.

For ZS 3D, we use point clouds alone to identify and localize anomalies. In ZS multimodal 3D, both point clouds and their corresponding 2D RGB counterparts are available. To comprehensively evaluate PointAD+ and PointAD, we adopt four metrics to evaluate its anomaly classification and segmentation performance. For global detection, we use the Area Under the Receiver Operating Characteristic Curve (I-AUROC$\%\uparrow$) and average precision (AP$\%\uparrow$). As for local anomaly segmentation, we use point-level AUROC (P-AUROC$\%\uparrow$) and a restricted metric called AUPRO$\%(\uparrow)$~\cite{bergmann2020uninformed} to provide a detailed evaluation of subtle anomaly regions.
\begin{table*}[]
 \caption{Performance comparison on ZS 3D anomaly detection in "one-vs-rest" setting. Note that RealAD-3D could not compute the AUPRO. The number in brackets next to the dataset name indicates its number of object categories.}
 \vspace{-0.5em}
\label{table: Performance comparison on ZS 3D anomaly detection in "one-vs-rest" setting.}
\setlength\tabcolsep{2pt} 
\centering
\begin{tabular}{cccccccc ccccccc}
\toprule
Dataset  & \multirow{2}{*}{\makecell[c]{Detec. \\level}} &\multicolumn{2}{c}{MVTec3D-AD(10)}  &  \multicolumn{2}{c}{Eyecandies(10)} & \multicolumn{1}{c}{Real3D-AD(12)}  & \multirow{2}{*}{\makecell[c]{Detec. \\level}} &\multicolumn{2}{c}{MVTec3D-AD(10)}  & \multicolumn{2}{c}{Eyecandies(10)} & \multicolumn{2}{c}{Real3D-AD(12)} \\  \cline{3-7}\cline{9-14}
Metric& &P-AUROC&AUPRO&P-AUROC&AUPRO&P-AUROC & &I-AUROC&AP&I-AUROC&AP&I-AUROC&AP \\ \hline 

CLIP + R.  & \multirow{9}{*}{\makecell[c]{L.}} & -& 54.4 & 81.2 & 37.9 & 45.9 & \multirow{9}{*}{\makecell[c]{G.}} 
&  61.2& 85.8 & 66.7& 69.2 & 68.8 & 72.3  \\

Cheraghian &   &88.2 & 57.0 & - & - & -  & & 53.6& 81.7& 49.5& 48.1 & 50.3 & 54.4   \\

PoinCLIP V2 &   & 87.4& 52.3 & 43.7 & - & 52.9  & & 51.2& 80.1 & 46.1& 48.1 &  53.1 & 58.1  \\

PointCLIP V2$_a$ &  &87.3& 52.3 & 44.2 & - & 52.2  & & 51.1 & 80.6 & 44.4 & 47.0 & 57.5 
  & 58.3
 \\

AnomalyCLIP &   &88.9& 60.9 & 77.7 & 43.4 & 50.3  & & 56.4& 83.5 & 57.6& 59.0 & 55.2 & 57.1 \\

\rowcolor{gray!40}
PointAD-CoOp &  &94.8 & 82.0 & 91.5 & \textcolor{blue}{71.3} & 72.6  & & 80.9 & 93.9 & 67.7 & 71.8 & 73.9 & 75.9 \\

\rowcolor{gray!40}
PointAD&  &95.5 & 84.4 & \textcolor{blue}{92.1} & \textcolor{blue}{71.3} & 73.5  & & 82.0 & 94.2 & \textcolor{blue}{69.1} & \textcolor{blue}{73.8} & 74.8 & 76.9\\

\rowcolor{gray!40}
PointAD+-CoOp&  &\textcolor{blue}{96.0} & \textcolor{blue}{87.2} & 91.9 & 70.6 & \textcolor{blue}{75.1}  & & \textcolor{blue}{82.3} & \textcolor{blue}{94.4} & 68.7 & 73.2 & \textcolor{blue}{81.6} & \textcolor{blue}{83.9} \\

\rowcolor{gray!40}
PointAD+&  &\textcolor{red}{96.9} & \textcolor{red}{88.7} & \textcolor{red}{92.6} & \textcolor{red}{72.1} & \textcolor{red}{76.0}  & & \textcolor{red}{82.5} & \textcolor{red}{94.8} & \textcolor{red}{69.9} & \textcolor{red}{74.2} & \textcolor{red}{83.5} & \textcolor{red}{84.7} \\
\bottomrule
\vspace{-2em}
\end{tabular}%
\end{table*}

\begin{table*}[]
 \caption{Performance comparison on ZS multimodal 3D anomaly detection in "one-vs-rest" setting.}
 \vspace{-0.5em}
\label{table: Performance comparison on ZS multimodal 3D anomaly detection in "one-vs-rest" setting.}
\setlength\tabcolsep{2pt} 
\centering
\begin{tabular}{ccccccccccc}
\toprule
Dataset & \multirow{2}{*}{\makecell[c]{Detec. \\level}} & \multicolumn{2}{c}{MVTec3D-AD(10)}  & \multicolumn{2}{c}{Eyecandies(10)} & \multirow{2}{*}{\makecell[c]{Detec. \\level}} &\multicolumn{2}{c}{MVTec3D-AD(10)}  & \multicolumn{2}{c}{Eyecandies(10)} \\  \cline{3-6}\cline{8-11}
Metric&&P-AUROC&AUPRO&P-AUROC&AUPRO & &I-AUROC&AP&I-AUROC&AP  \\ \hline 

CLIP + R. &\multirow{9}{*}{\makecell[c]{ML.}}  &- & 56.0 & 78.0& 31.8 & \multirow{9}{*}{\makecell[c]{MG.}}  & 60.4& 86.4 & 73.0 & 73.9 \\

Cheraghian&   &-& - &-& - & &-&-&-& - \\

PoinCLIP V2&    &78.3& 49.4 &46.0& - & &49.8 & 79.3 & 46.9 & 49.9 \\

PointCLIP V2$_a$&  &79.5 &51.6 & 46.2 & - & & 49.4 & 79.8 & 48.5 & 50.5 \\

AnomalyCLIP&   &91.6 & 70.9 &85.0& 56.2 & & 66.2 & 87.6 & 65.0 & 67.5 \\

\rowcolor{gray!40}
PointAD-CoOp&  & 96.5 & 88.8 & 94.9 & \textcolor{blue}{83.6} & & 83.4 & 94.9 & 73.7 & 76.0  \\

\rowcolor{gray!40}
PointAD&  & 97.2 & 90.2 & \textcolor{red}{95.3} & \textcolor{red}{84.3} & &86.9 & \textcolor{blue}{96.1} & 77.7 & \textcolor{blue}{80.4}\\

\rowcolor{gray!40}
PointAD+-CoOp&  & \textcolor{blue}{97.8} & \textcolor{blue}{92.6} & 95.0 & 83.5 & & \textcolor{blue}{87.5} & \textcolor{blue}{96.1} & \textcolor{blue}{78.0} & 79.8 \\

\rowcolor{gray!40}
PointAD+&  & \textcolor{red}{98.2} & \textcolor{red}{93.1} & \textcolor{blue}{95.2} & \textcolor{red}{84.3} & & \textcolor{red}{87.7} & \textcolor{red}{96.6} & \textcolor{red}{79.2} & \textcolor{red}{81.7}\\

\bottomrule
\end{tabular}%
\end{table*}

\textbf{Implementation details.} Both point clouds and 2D renderings are resized to 336 $\times$ 336. We use Open3D library to generate 9 views by rotating point clouds along the X-axis at angles of $\{-\frac{4}{5}\pi, -\frac{3}{5}\pi, -\frac{2}{5}\pi, -\frac{1}{5}\pi, 0, \frac{1}{5}\pi, \frac{2}{5}\pi, \frac{3}{5}\pi, \frac{4}{5}\pi \}$ for most categories. Some objects lose their surface when rendered at certain angles, we manually adjust them to an appropriate angle. The backbone of PointAD+ and PointAD is the pre-trained CLIP model (\texttt{VIT-L/14@336px} in \texttt{open\_clip}). Following~\cite{zhou2023anomalyclip}, we improve the local visual semantics of the vision encoder of CLIP without modifying its parameters. During training, we keep all parameters of CLIP frozen and set the learnable word embeddings in object-agnostic text templates to $12$. The hyperparameters, including the number of neighbors $k$, the weighting scalar $\alpha$, and the scale parameter $\sigma$ of the Gaussian kernel, are set to 10, 0.5, and 1, respectively. We use the Adam optimizer with a learning rate of 0.001 to optimize the learnable parameters. The experiment runs for 15 epochs with a batch size of 4. All experiments were conducted on a single NVIDIA RTX 3090 24GB GPU using PyTorch-2.0.0. PointAD+ and PointAD by default use object-agnostic text prompts, whereas PointAD+-CoOp and PointAD-CoOp employ object-aware prompts.

\textbf{Baselines.} As ZS 3D and multimodal 3D anomaly detection remain underexplored areas in the field, we make a great effort to provide these comparisons. We apply the original CLIP to our framework for 3D detection, called CLIP + Rendering. Also, we reproduce SOTA 3D recognition works including PointCLIP V2~\cite{zhu2023pointclip} and Cheraghian~\cite{cheraghian2022zero} and adapt them for ZS 3D anomaly detection. We compare the SOTA 2D anomaly detection approach AnomalyCLIP~\cite{zhou2023anomalyclip} by fine-tuning it on depth maps. All hyperparameters in these baselines are kept the same. We will present the detailed reproduction:
\begin{itemize}
    \item CLIP + Rendering is a method, where we apply the original CLIP into our framework for ZS 3D anomaly detection. It uses the same rendering procedure as PointAD+ and PointAD. Following~\cite{jeong2023winclip, zhou2023anomalyclip}, we integrate anomaly semantics into CLIP by two class text prompt templates: \verb'A photo of a normal [cls]' and \verb'A photo of an anomalous [cls]', where \verb'cls' denotes the target class name.
    \item PointCLIP V2 (CVPR 2023) is a SOTA ZS 3D classification method based on CLIP. They project point clouds into depth maps from different views. To adapt PointCLIP V2 into ZS anomaly detection, we replace its original text prompts~\texttt{point cloud of a big [c]} with normal text prompts~\texttt{point cloud of a big [c]} and abnormal text prompts~\texttt{point cloud of a big damaged [c]}. PointCLIP V2$_a$ is its adaptation version, fine-tuned using the same data we employed.
    \item AnomalyCLIP (ICLR 2024) is a SOTA ZS 2D anomaly detection method. AnomalyCLIP introduces object-agnostic learning to capture generic anomaly semantics of images. We adapt AnomalyCLIP in 3D detection by fine-tuning AnomalyCLIP on depth images.
    \item Cheraghian (IJCV 2022) is an approach for ZS 3D classification without foundation models. They directly extract the point presentations by PointNet and use word2vector~\cite{mikolov2013efficient} to generate the textual embedding of \texttt{[c]}. To incorporate the anomaly semantics into Cheraghian, we average the textual embeddings of \texttt{[c]} and \texttt{damaged}. We replace the global representation with dense representations to provide the segmentation.
\end{itemize}

\begin{figure*}[h]
  \begin{minipage}{0.58\textwidth}
        \centering

 \captionof{table}{Performance comparison on ZS 3D anomaly detection in cross-dataset setting. Note that RealAD-3D could not compute the AUPRO.}
 \vspace{-0.5em}
\label{table: Performance comparison on ZS 3D
anomaly detection in across-dataset setting.}
\setlength\tabcolsep{0.5pt} 
\footnotesize
\begin{tabular}{ccccc ccccc}
\toprule
 Dataset & \multirow{2}{*}{\makecell[c]{Detec. \\level}}& \multicolumn{2}{c}{Eyecandies} & \multicolumn{1}{c}{Real3D-AD} & \multirow{2}{*}{\makecell[c]{Detec. \\level}}  & \multicolumn{2}{c}{Eyecandies} & \multicolumn{2}{c}{Real3D-AD}  \\  \cline{3-5}\cline{7-10}
Metric && AUROC&PRO&AUROC & &I-AUROC&AP&I-AUROC&AP \\
\hline
 
PointCLIP V2$_a$ & \multirow{2}{*}{\makecell[c]{L.}} & 43.9 & - & 51.9  & \multirow{2}{*}{\makecell[c]{G.}} & 45.2 & 48.0 & 57.4 & 58.8  \\

AnomalyCLIP&  & 45.4 & 50.3 & 14.8 & & 56.3 & 57.1 & 52.7 & 55.7 \\

\rowcolor{gray!40}
PointAD-CoOp&  & 91.8 & 70.5 & 70.1  & & 69.1 &73.8 & \textcolor{blue}{74.8} & 76.9\\

\rowcolor{gray!40}
PointAD&  & 91.8 & 71.4 & \textcolor{blue}{71.6}  & & \textcolor{blue}{69.5} & \textcolor{blue}{74.3} & \textcolor{red}{75.9} & \textcolor{blue}{77.9} \\

\rowcolor{gray!40}
PointAD+-CoOp&  & \textcolor{blue}{92.5} & \textcolor{blue}{72.1} & \textcolor{blue}{71.6}  & & 69.0 & 74.1 & 74.0 & 76.6 \\

\rowcolor{gray!40}
PointAD+&  & \textcolor{red}{92.7} & \textcolor{red}{72.8} & \textcolor{red}{72.5}  & & \textcolor{red}{70.2} & \textcolor{red}{74.6} & 74.3 & \textcolor{red}{78.2}\\
\bottomrule
\end{tabular}%
    \end{minipage} 
    \hfill
    \hfill
\begin{minipage}{0.4\textwidth}
    \centering
    \captionof{table}{Performance on ZS multimodal 3D
anomaly detection in cross-dataset setting.}%
\vspace{-0.5em}
\label{table: Performance comparison on ZS multimodal 3D
anomaly detection in across-dataset setting.}
\setlength\tabcolsep{0.5pt} 
\footnotesize
\begin{tabular}{cccc ccc}
\toprule
Dataset & \multirow{2}{*}{\makecell[c]{Detec. \\level}}&  \multicolumn{2}{c}{Eyecandies} & \multirow{2}{*}{\makecell[c]{Detec. \\level}} & \multicolumn{2}{c}{Eyecandies}  \\  \cline{3-4} \cline{6-7}
Metric& &AUROC&PRO & &I-AUROC&AP\\
\hline

PointCLIP V2$_a$ & \multirow{2}{*}{\makecell[c]{ML.}} &46.3 & - & \multirow{2}{*}{\makecell[c]{MG.}}& 48.5 & 50.9 \\

AnomalyCLIP&   & 86.2& 61.3 & & 65.7& 68.1  \\

\rowcolor{gray!40}
PointAD-CoOp &  & 94.4 & 80.3 & & 76.3 & 78.9 \\

\rowcolor{gray!40}
PointAD&  & 94.0 & 80.7 & & 78.6 & 80.8\\

\rowcolor{gray!40}
PointAD+-CoOp&  & \textcolor{blue}{94.5} & \textcolor{blue}{84.8} & & \textcolor{blue}{79.8} & \textcolor{red}{83.5}\\

\rowcolor{gray!40}
PointAD+&  & \textcolor{red}{94.8} & \textcolor{red}{85.2} & & \textcolor{red}{80.3} & \textcolor{blue}{83.1}\\
\bottomrule
\end{tabular}%
    \end{minipage} 
\end{figure*}

\begin{figure*}[t]
  \centering
    \subfigure[Multimodal visualization with hybrid loss.]
    {\includegraphics[width=0.42\textwidth]{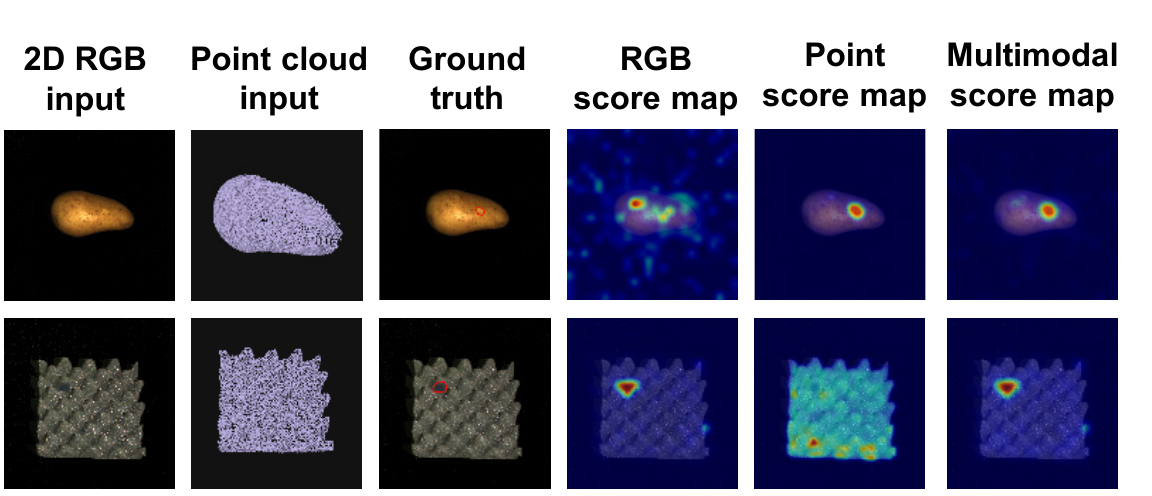}%
    \label{fig6: hybrid loss}}
   \hfil 
    \subfigure[Multimodal visualization without 2D glocal loss.]
    {\includegraphics[width=0.42\textwidth]{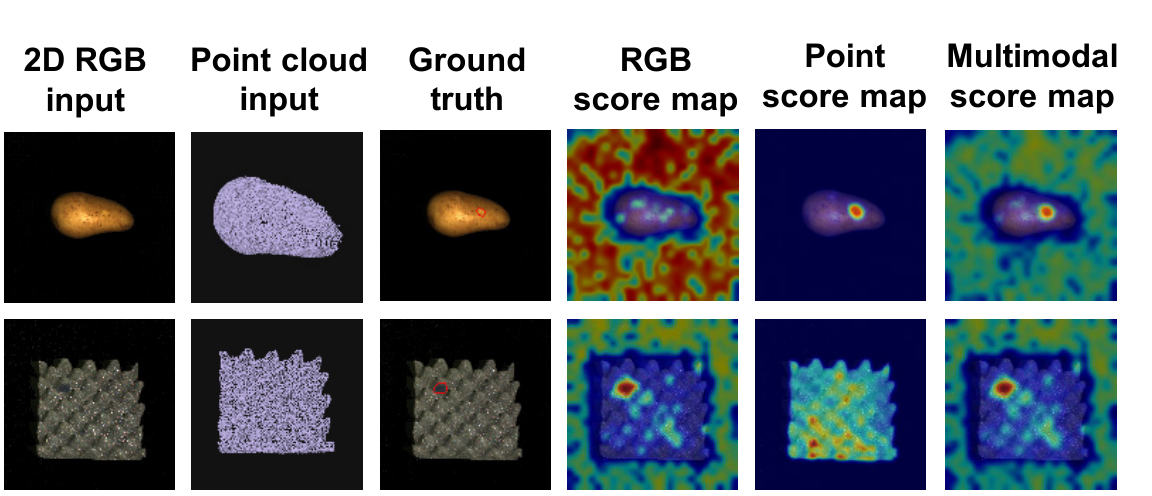}%
    \label{fig6: hybrid loss without 2D glocal loss}}
 \caption{Visualization comparison on ZS multimodal 3D between (a) with hybrid loss and (b) without 2D glocal loss.}
 \label{fig:auroc_curves}
\end{figure*}
\subsection{Main Results}
\label{main results}
We train PointAD and PointAD+ on three objects on MVTec3D-AD, Eyecandies, and Real3D-AD. Over three runs, the averaged results on \textbf{one-vs-rest} and \textbf{cross-dataset} settings are reported. We use metric pairs (I-AUROC\% and AP\%) and (P-AUROC\% and AUPRO\%) to evaluate the global and local detection performance, respectively. The best and second-best results in ZS are highlighted in~\textcolor{red}{Red} and~\textcolor{blue}{Blue}. G. and L. represent 3D global and local anomaly detection. Multimodal 3D global and local detection are abbreviated as MG. and ML.

\textbf{ZS 3D anomaly detection.} Table~\ref{table: Performance comparison on ZS 3D
anomaly detection in "one-vs-rest" setting.} presents the comparison of ZS 3D performance. Compared to the point-based method Cheraghian and the projection-based method PointCLIP V2, PointAD+ and PointAD achieve superior performance on ZS 3D anomaly detection over all three datasets across 32 object semantics. Especially, PointAD outperforms CLIP + Rendering from 61.2\% to 82.0\% I-AUROC and from 85.8\% to 94.2\% AP on MVTec3D-AD. PointAD+ further obtains the performance gain over PointAD. Besides global anomaly detection, PointAD+ and PointAD also attain superior segmentation performance. PointAD improves MVTec3D-AD by a large margin compared to Cheraghian from 88.2\% to 95.5\% P-AUROC and from 57.0\% to 84.4\% AUPRO. This improvement in overall performance is attributed to PointAD adapting CLIP's strong generalization to glocal anomaly semantics through hybrid representation learning. On this basis, PointAD+ further improves the performance from 95.5\% to 96.8\% P-AUROC and from 84.4\% to 88.4\% AUPRO. It is attributed to the hierarchy representation learning, incorporating explicit point relations to better reveal the abnormality within point clouds. As shown in Figure~\ref{f5:visualization}, PointAD+ and PointAD could accurately segment the unseen anomalies across diverse objects. In addition, PointAD+ and PointAD advance PointAD+-CoOp and PointAD-CoOp across all datasets by blocking the class semantics in text prompts~\cite{zhou2023anomalyclip}.

\textbf{ZS multimodal 3D anomaly detection.} PointAD+ and PointAD allow the direct incorporation of RGB counterpart when it is available during the test. Table~\ref{table: Performance comparison on ZS multimodal 3D
anomaly detection in "one-vs-rest" setting.} indicates that PointAD+ and PointAD leverage additional RGB information in a plug-and-plug manner to boost their performance. PointAD gets the performance gain from 82.0\% to 86.9\% AUROC and from 94.2\% to 96.1\% AP for global semantics on MVTec3D-AD. As for local semantics, the performance improves from 95.5\% to 97.2\% P-AUROC and from 84.4\% to 90.2\% AUPRO. A large performance gain is also obtained on Eyecandies and Real3D-AD. PointAD+ can also benefit from the RGB information and break the performance achieved by PointAD, i.e., 98.2\% vs. 97.2\% P-AUROC and 93.1\% vs. 90.2\% AUPRO on MVTec3D-AD, and 79.2\% vs. 77.7\% I-AUROC and 81.7\% vs. 80.4\% AP on Eyecandies. As other methods can improve their performance in some metrics, they still suffer from performance degradation in other metrics due to inefficient integration of the two modalities. Instead, PointAD+ achieves overall improvement by incorporating explicit joint constraints on both point and pixel information.

\textbf{Cross-dataset ZS anomaly detection.} Cross-dataset detection is conducted to further evaluate the ZS capacity of PointAD+ and PointAD, using one object as an auxiliary and testing objects with totally different semantics and scenes in another dataset. We compare all baselines that need fine-tuning in Table~\ref{table: Performance comparison on ZS
3D anomaly detection in across-dataset
setting.} and multimodal 3D from Table~\ref{table: Performance comparison on ZS
multimodal 3D anomaly detection in across-dataset
setting.}. PointAD demonstrates strong cross-dataset generalization performance on Eyecandies and Real3D-AD, with nearly no obvious performance decay compared to the one-vs-rest setting. The strong transfer ability highlights its robust generalization capabilities in detecting anomalies in objects with unseen semantics and backgrounds. Compared to PointAD, PointAD+ exhibits a performance drop on Real3D-AD, with P-AUROC decreasing from 76.0\% to 72.5\%, I-AUROC dropping from 83.5\% to 74.3\%, and AP declining from 84.7\% to 78.2\%. The main advantage of PointAD+ over PointAD is to introduce explicit point learning and reveal the point relations. Hence, its performance gain depends on the gap in geometry between the source and target point clouds. Despite this, PointAD+ generally achieves comparable or superior performance to PointAD. We would like to point out that developing a more generalized point encoder to model point relations could help mitigate this spatial gap, thereby further improving its ZS detection capacity. We leave it for future research.

\vspace{-0.8em}
\subsection{Result Analysis}
PointAD+ characterizes itself by capturing abnormality from two aspects: implicit 3D abnormality, where point anomalies are projected into the pixel space; explicit 3D abnormality, which reveals the inherent point relations. In this section, we provide a more in-depth analysis of how they contribute to the success of PointAD+.

\begin{figure}[t]
    \centering
    \includegraphics[width=1\linewidth]{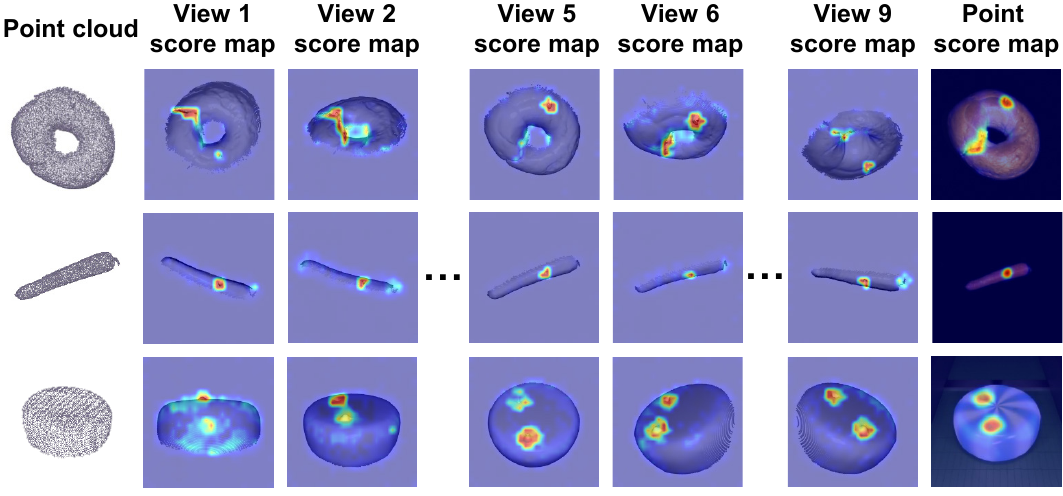}
    \caption{Visualization on ZS 3D. Each row visualizes the anomaly score maps of 2D renderings from different views, and the final point score maps are also presented.}
    \vspace{-0.5em}
    \label{fig:enter-label}
\end{figure}

\begin{figure}
    \centering
    \includegraphics[width=1\linewidth]{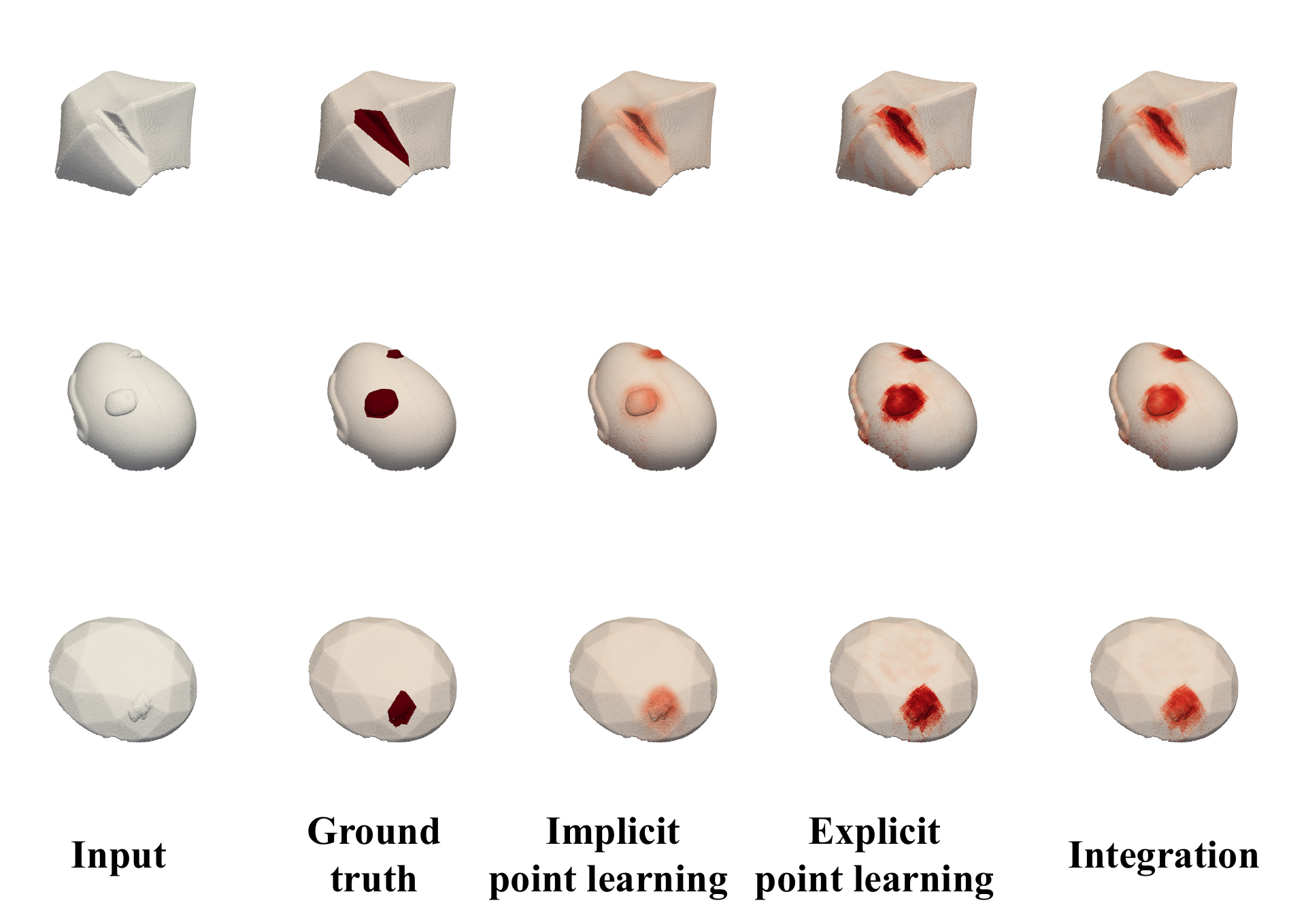}
    \caption{\textcolor{orange}{Visualization on stereo point clouds (Real3D-AD)}}
    \label{fig:realad-vis}
\end{figure}
\subsubsection{How multimodality makes PointAD+ accurate.}
PointAD+ and PointAD are unified frameworks that can not only capture point anomalies but also handle 2D information in a plug-and-play manner. To study how it works, we visualize the multimodal 3D results of PointAD on MVTec3D-AD in Figure~\ref{fig6: hybrid loss}. The surface damage on the potato presents a similar appearance to the object foreground, which makes it difficult to detect this anomaly with RGB information. On the contrary, the point relations for the color stain on foam are the same as those of normal, but they have a clear distinction in the RGB information. PointAD can integrate these two modalities, thereby complementing their respective advantages. We further investigate the reason why PointAD+ and PointAD can directly leverage both modalities. For this purpose, we experiment without 2D glocal loss. As shown in Figure~\ref{fig6: hybrid loss without 2D glocal loss}, without 2D glocal loss, significant noise disrupts and even covers the RGB score maps, resulting in unpromising multimodal fusions. This illustrates the importance of explicit constraints on the 2D space. Hence, we conclude that the robust multimodal detection capability of our model stems from the collaboration optimization in both 3D and 2D spaces during training.

\subsubsection{Implicit point learning on rendering layer} We first investigate how rendering anomalies facilitate 3D anomaly detection. As illustrated in Figure~\ref{fig:enter-label}, we visualize the anomaly score maps of the 3D and corresponding 2D counterparts of PointAD+ on MVTec3D-AD. And, we also visualize the anomaly score map of 2D counterparts according to Eq.~\ref{equ:rendering2d}, projected by 3D point anomalies. It can be observed that PointAD+ has a strong detection ability on such 2D anomalies. The strong representative pixel representations from multiple views facilitate more precise 3D anomaly detection. We conduct a quantitative analysis of the view number in the ablation study. The strong 3D and 2D detection capabilities of PointAD+ are from hybrid representation learning, which not only enables PointAD+ to capture the 3D anomalies but also explicitly constrains 2D representations.

\subsubsection{Explicit point learning on geometric layer}
We further analyze what PointAD+ learns from the geometry layer as defined in Eq.~\ref{equal:geometry3d}. As illustrated in Figure~\ref{fig:realad-vis}, the explicit point learning module excels at capturing abnormalities in stereo point clouds, as their abnormalities rely heavily on underlying spatial relations. By incorporating geometric information, it exhibits a stronger capability to detect abnormal point relations, outperforming implicit point learning in identifying such structural anomalies. However, explicit point learning may introduce false positives near object boundaries in the point cloud. This observation is consistent with our discussion in Section~\ref{sec:Motivation of PointAD+}.

\begin{figure}[t]
  \begin{minipage}{0.45\textwidth}
        \centering
 \captionof{table}{ \textcolor{orange}{Comparison of computation overhead with SOTA approaches on MVTec3D-AD.}}
 \vspace{-0.5em}
 \label{table Ablation on Complexity analysis.}
 \scriptsize
    \setlength\tabcolsep{1pt} 
\begin{tabular}{c|cccccc}
    \toprule
    \multirow{2}{*}{Methods} & \multirow{2}{*}{\makecell[c]{GFLOPs}} & \multirow{2}{*}{\makecell[c]{FPS}} & \multirow{2}{*}{\makecell[c]{GPU memory\\ usage (Peak)}} & \multicolumn{2}{c}{Point detection}  \\ 
    &&&& Local   & Global   \\ \hline
    PointCLIP V2 & 4660.49& 0.66 & 9747MB& (87.4, 52.3) & (78.3, 49.4)  \\
    CLIP + Rendering & 1057.36 & 3.61 & 3685MB & (61.2, 85.8) & (-, 54.4)\\  
    Cheraghian & 31.49&  2.86 & 4847MB&  (88.2, 57.0) & (53.6, 81.7) \\   
     AnomalyCLIP & 1656.34 & 5.26 & 3348MB& (88.9, 60.9) & (56.4, 83.5) \\  
    \rowcolor{gray!40}
    PointAD &  1656.34 & 2.52& 4275MB & (\textcolor{blue}{95.5}, \textcolor{blue}{84.4}) & (\textcolor{blue}{82.0}, \textcolor{blue}{94.2})\\
    \rowcolor{gray!40}
    PointAD+ &  1658.34 & 2.18 & 5811MB & (\textcolor{red}{96.9}, \textcolor{red}{88.7}) & (\textcolor{red}{82.5}, \textcolor{red}{94.8}) \\
    \bottomrule
    \end{tabular}
    \end{minipage}
\end{figure}

\subsubsection{Complexity analysis}
\label{Complexity analysis}
The computation overhead is also an indicator to assess the effectiveness of the proposed method. We compare the computation overhead in Table~\ref{table Ablation on Complexity analysis.}, including GFLOPs, frames per second (FPS), and GPU memory consumption with a batch size of 1.  \textcolor{orange}{For a fair comparison, we maintain NVIDIA RTX 3090 24GB GPU idle. PointAD+ requires only about 2.0 more GFLOPs than PointAD (1658.34 vs. 1656.34) yet achieves a significant performance improvement. In contrast, other CLIP-based methods, such as PointCLIP V2, consume larger GFLOPs but deliver worse results. Although CLIP+Rendering, Cheraghian, and AnomalyCLIP use fewer GFLOPs, they fail to provide competitive performance.} As for inference speed, compared to PointCLIP V2, PointAD requires less time to infer an image and achieves a higher 2.52 vs. 0.66 FPS with lower graphic memory usage. PointAD+ needs a slightly higher computation overhead over PointAD, i.e., 2.52 vs. 2.18 FPS, and attains a noticeable performance improvement on ZS 3D and multimodal 3D. CLIP + rendering has the fastest inference speed, but its detection performance lags behind PointAD+ and PointAD. From the above results, PointAD+ achieves a favorable trade-off between performance and overhead.

\section{Ablation Study}
In this section, we analyze the impact of the proposed modules and examine the sensitivity of key hyperparameters.

\begin{figure}[t]
  \hspace{+0.0cm}\begin{minipage}{0.53\textwidth}
        \centering
 \captionof{table}{Ablation on proposed modules. F stage and S stage represent the first and second stages of G-aggregation, respectively.}
 \vspace{-0.5em}
 \label{table: Ablation on the proposed modules.}
\setlength\tabcolsep{1pt}
 \scriptsize
    \begin{tabular}{cccc|ccccccc}
    \toprule
\multicolumn{4}{c|}{$L_{hybrid}^r$} & \multicolumn{2}{c}{$L^{geo}$} & \multirow{2}{*}{\makecell[c]{$L^{c}$}} & \multicolumn{2}{c}{Point detection} & \multicolumn{2}{c}{Multimodal detection} \\
      $L_{I(3d)}^{global}$ &$ L_{I(3d)}^{local}$ & $L_{I(2d)}^{global}$ & $L_{I(2d)}^{local}$ & F & S & & Local & Global & Local & Global \\
    \midrule
        
    & & & &    &  &   & (-, 54.4) & (61.2, 85.8)  & (-, 56.0) & (60.4, 86.4)\\

    \ding{51}& &  &  &   &  &    & (91.9, 71.7)& (75.5, 91.9)  & (92.6, 81.6)& (80.4, 93.9)\\
    \ding{51}& \ding{51} & &  &   &  &    & (95.2, 82.7)& (81.3, 94.1)  & (92.0, 81.4)& (83.9, 95.0)\\

   &  & \ding{51} &  \ding{51}&   &  &  &  (93.9, 82.8)&	(79.3, 91.6)	&(91.0, 82.2)	&(82.6, 94.1)\\

     & \ding{51} &\ding{51} & \ding{51} &   &  &   & (95.5, 84.7)&	(81.8, 92.3) &(96.1, 90.6) &	(83.7, 95.1)\\
                     
     \ding{51}&\ding{51} & \ding{51} & &   &  &    & (95.6, 82.5)& (82.4, 94.5)  & (92.9, 84.4)& (85.5, 95.6)\\
     \ding{51}& \ding{51} & \ding{51} & \ding{51}&   &  &    & (95.5, 84.4)& (82.0, 94.2)  & (97.2, 90.2)& (86.9, 96.1)\\

     \ding{51}& \ding{51} & \ding{51} & \ding{51}& \ding{51}  &  &   &  (95.9, 87.1)& (82.2, 94.4)  & (97.0, 91.4)& (87.6, 96.3)\\
  
     \ding{51}& \ding{51} & \ding{51} & \ding{51}&  \ding{51} & \ding{51} &   &  (96.3, 87.5)& (81.5, 94.1)  & (97.9, 92.5)& (87.1, 96.4)\\

    \textcolor{orange}{ 
     \ding{51}} & \textcolor{orange}{ \ding{51} }& \textcolor{orange}{  \ding{51} } & \textcolor{orange}{  \ding{51} } & \textcolor{orange}{  \ding{51} } &  & \textcolor{orange}{  \ding{51} } & \textcolor{orange}{  (96.2, 88.0) }& \textcolor{orange}{(82.6, 94.7) } & \textcolor{orange}{  (97.5, 92.1) }& \textcolor{orange}{ (87.8, 96.7)}
     \\

    \textcolor{orange}{ 
     \ding{51}} & \textcolor{orange}{ \ding{51} }& \textcolor{orange}{  \ding{51} } & \textcolor{orange}{  \ding{51} } &  &
     \textcolor{orange}{  \ding{51} } & \textcolor{orange}{  \ding{51} } & \textcolor{orange}{  (96.6, 88.4) }& \textcolor{orange}{(82.1, 94.3) } & \textcolor{orange}{  (98.0, 92.7) }& \textcolor{orange}{ (87.4, 96.6)}
     \\
     
    \rowcolor{gray!40}
     \ding{51}& \ding{51} & \ding{51} & \ding{51}&  \ding{51} & \ding{51} & \ding{51}  &  (\textcolor{red}{96.9}, \textcolor{red}{88.7})& (\textcolor{blue}{82.5}, \textcolor{red}{94.8})  & (\textcolor{red}{98.2}, \textcolor{red}{93.1})& (\textcolor{red}{87.7}, \textcolor{blue}{96.6})\\
    \bottomrule
\end{tabular}
    \end{minipage} 
\end{figure}

\subsection{Ablation on modules and key hyperparameters}
\textbf{Module ablation.} We evaluate the effectiveness of the proposed techniques by progressively incorporating the introduced modules. As shown in Table~\ref{table: Ablation on the proposed modules.}, the baseline \textit{vanilla} model, which represents CLIP + rendering, performs poorly on both 3D and multimodal 3D anomaly detection due to CLIP's focus on aligning 2D object semantics rather than anomaly semantics. Adding the 3D global branch incorporates global anomaly semantics into PointAD+, improving both local and global anomaly detection performance. Although the 3D local branch further enhances performance, local accuracy on multimodal 3D detection declines due to the absence of 2D constraints, leading to inefficient multimodal fusion when integrating 2D RGB information in a plug-and-play manner. The inclusion of the 2D global branch explicitly introduces 2D anomaly information, enabling PointAD+ to achieve overall performance gains and better multimodal fusion. By focusing on 2D anomaly regions, PointAD+ develops a deeper understanding of point clouds from 2D representations, improving multimodal segmentation performance from 92.9\% to 97.2\% P-AUROC and from 84.4\% to 90.2\% AUPRO. So far, PointAD+ does not explicitly account for point relationships within point clouds. To address this, explicit point learning is introduced to model abnormal spatial relationships. PointAD+ leverages geometric characteristics, achieving notable improvements in segmentation performance. Specifically, P-AUROC increases from 95.5\% to 96.3\% and AUPRO rises from 84.4\% to 87.5\% on ZS 3D detection. On multimodal 3D detection, P-AUROC improves from 97.2\% to 97.9\% and AUPRO increases from 90.2\% to 92.5\%. Finally, cross-alignment is introduced to enhance interactions between hierarchical representations, leading to further performance gains over these metrics.

 \textcolor{orange}{
\textbf{Ablation on the number of neighbors}
The number of neighbors $k$ determines the receptive field of each point within its local neighborhood. As illustrated in Table~\ref{table Ablation on neighbors.}, we further explore $k \in \{5, 10, 15, 20, 25\}$ to study its impact. For point detection, the best local accuracy is achieved at $k=10$ (96.9 AUROC and 88.7 AUPRO), and the best global accuracy is also observed at $k=10$ (82.5 AUROC and 94.8 AP). For multimodal detection, both $k=10$ and $k=20$ yield superior performance. Overall, the detection results remain stable across different values of $k$. This indicates that PointAD+ is robust to the number of neighboring points.
}

\begin{figure}[h]
\hfill
  \hspace{+0.8cm}\begin{minipage}{0.45\textwidth}
        \centering
 \captionof{table}{ \textcolor{orange}{Ablation on the neighbors $k$.}}
 \vspace{-0.5em}
 \label{table Ablation on neighbors.}
 \scriptsize
    \setlength\tabcolsep{2pt} 
    \begin{tabular}{c|cccccc}
    \toprule
    \multirow{2}{*}{\makecell[c]{k}} & \multicolumn{2}{c}{Point detection} & \multicolumn{2}{c}{Multimodal detection} \\  &  Local & Global & Local  & Global  \\ \hline
     5  &  (96.5, \textcolor{blue}{88.6})& (82.4, 94.4)  & (\textcolor{blue}{98.0}, 93.0)& (87.5, 96.1)\\
    \rowcolor{gray!40}
     10  &  (\textcolor{blue}{96.9}, \textcolor{red}{88.7})& (\textcolor{blue}{82.5}, \textcolor{red}{94.8})  & (\textcolor{red}{98.2}, 93.1)& (\textcolor{blue}{87.7}, \textcolor{red}{96.6})\\
     15 & (\textcolor{red}{97.0}, 88.3) & (\textcolor{red}{82.6}, 94.4) & (97.9, \textcolor{blue}{93.2}) & (\textcolor{red}{87.8}, 96.4)\\
     20  &  (\textcolor{blue}{96.9}, 88.5)& (\textcolor{blue}{82.5}, \textcolor{blue}{94.5})  & (97.8, \textcolor{red}{93.5})& (87.6, \textcolor{blue}{96.5})\\ 
     25  &  (\textcolor{blue}{96.9}, 88.4)& (\textcolor{blue}{82.5}, \textcolor{blue}{94.5})  & (97.8, 93.1)& (87.6, 96.4)\\ 
    \bottomrule
    \end{tabular}
    \end{minipage}
\end{figure}

\begin{figure}[t]
  \hspace{+0.8cm}\begin{minipage}{0.45\textwidth}
        \centering
 \captionof{table}{Analysis on aggregation strategy $\ell$.}
 \vspace{-0.5em}
 \label{table Ablation on aggregation strategy.}
 \scriptsize
    \setlength\tabcolsep{2pt} 
    \begin{tabular}{c|cccccc}
    \toprule
    \multirow{2}{*}{\makecell[c]{$\ell$}} & \multicolumn{2}{c}{Point detection} & \multicolumn{2}{c}{Multimodal detection} \\  &  Local & Global & Local  & Global  \\ \hline
    \rowcolor{gray!40}
     Mean  &  (96.9, \textcolor{red}{88.7})& (\textcolor{red}{82.5}, \textcolor{red}{94.8})  & (\textcolor{red}{98.2}, 93.1)& (87.7, \textcolor{red}{96.6})\\
     Max & (\textcolor{blue}{97.0}, \textcolor{blue}{88.3}) & (\textcolor{red}{82.5}, \textcolor{blue}{94.4}) & (97.9, \textcolor{red}{93.5}) & (\textcolor{blue}{87.8}, 96.4)\\
    Parameteric network & (\textcolor{red}{97.1}, \textcolor{red}{88.7}) & (\textcolor{blue}{82.4}, \textcolor{blue}{94.4}) & (\textcolor{blue}{98.0}, \textcolor{blue}{93.2}) & (\textcolor{red}{88.0}, \textcolor{blue}{96.5})\\
    \bottomrule
    \end{tabular}
    \end{minipage}
\end{figure}

\begin{figure}[b]
    \centering
    \vspace{-1.6em}
    \subfigure[Local AUROC.]
   {\includegraphics[width=0.23\textwidth]{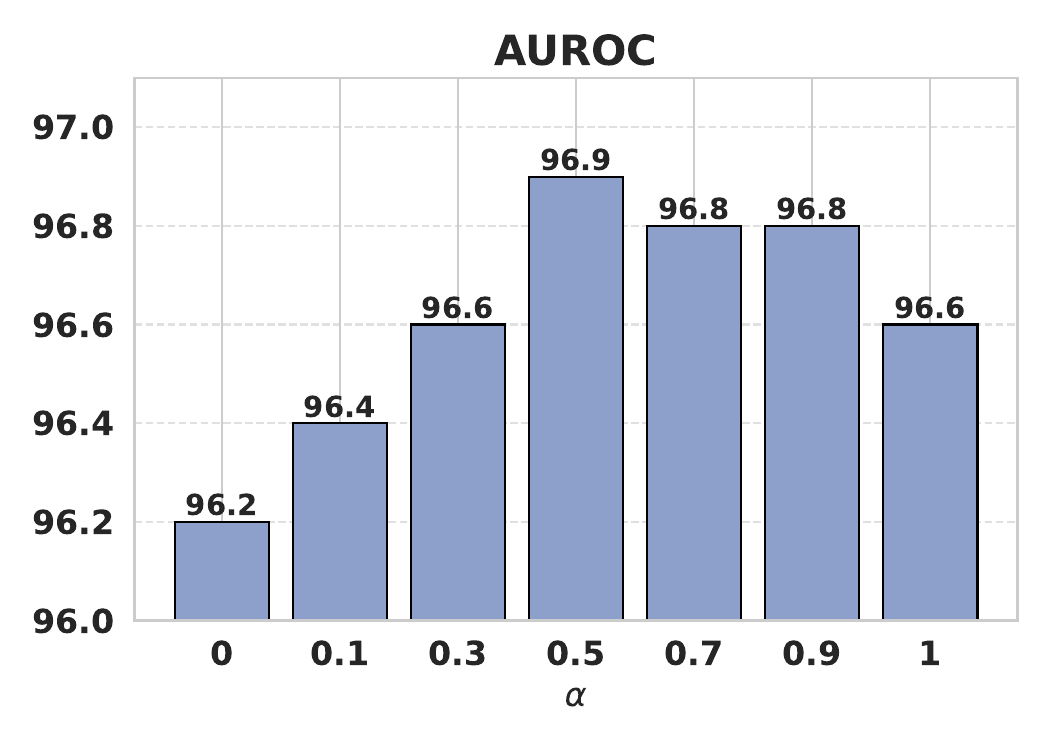}%
   \label{fig:lambda_3}
    }
        \hfil
               \hfil
  \subfigure[Multimodal Local AUROC.]
{\includegraphics[width=0.23\textwidth]{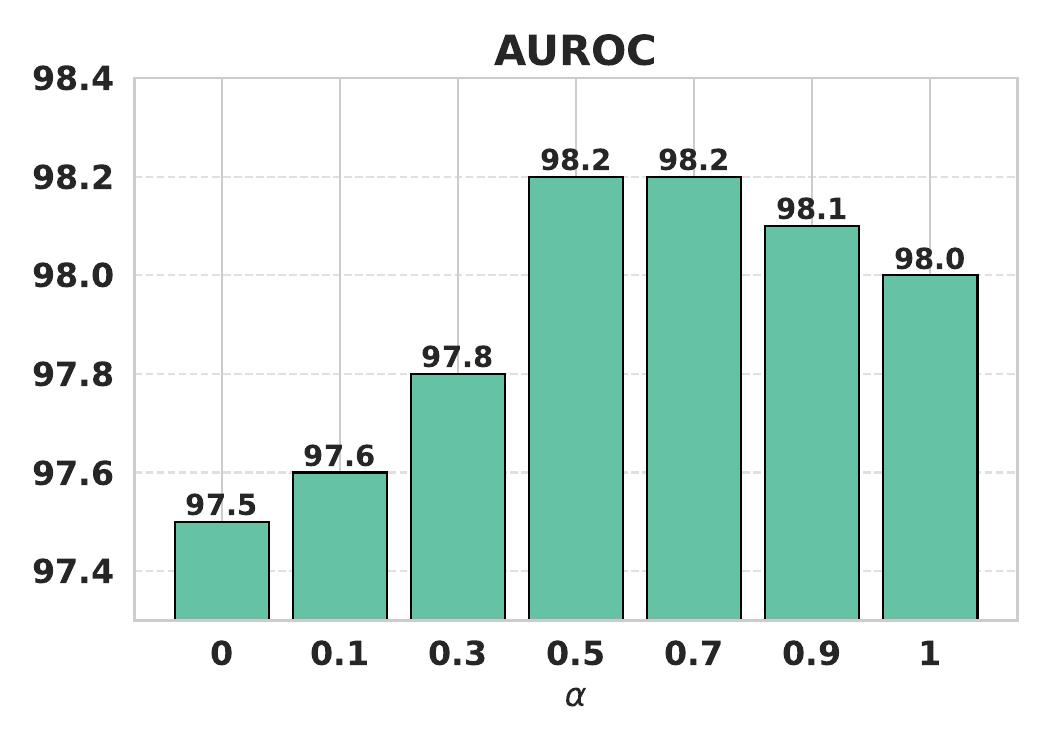}%
    \label{fig:lambda_1}}
       \caption{ \textcolor{orange}{The effect of $\alpha$ to the segmentation performance.}}
    \label{fig: hyperparameter Ablation}
\end{figure}

\begin{figure}[b]
\begin{minipage}{0.48\textwidth}

    \centering
    \captionof{table}{Ablation on the view number. The absolute numbers report PointAD+ performance. 
Arrows denote relative changes against baseline PointAD:  \textcolor{green!50!black}{↑} for improvements, purple \textcolor{purple}{↓} for degradations.}%
    \vspace{-0.5em}
    \label{table: Ablation on the view number.}
    \scriptsize
    \setlength\tabcolsep{2pt} 
        \begin{tabular}{ccccc}
        \toprule
        \multirow{2}{*}{\makecell[c]{K}} & \multicolumn{2}{c}{Point detection} & \multicolumn{2}{c}{Multimodal detection} \\
         & Local & Global & Local & Global \\
        \midrule
        1  & (95.8{\tiny\color{green!50!black}$\uparrow$1.7}, 83.7{\tiny\color{green!50!black}$\uparrow$4.6})
           & (74.0{\tiny\color{green!50!black}$\uparrow$1.4}, 90.7{\tiny\color{green!50!black}$\uparrow$0.6})
           & (97.3{\tiny\color{green!50!black}$\uparrow$1.3}, 90.6{\tiny\color{green!50!black}$\uparrow$3.0})
           & (81.3{\tiny\color{green!50!black}$\uparrow$0.9}, 94.2{\tiny\color{green!50!black}$\uparrow$0.3})\\
        3  & (96.4{\tiny\color{green!50!black}$\uparrow$1.2}, 86.4{\tiny\color{green!50!black}$\uparrow$3.9})
           & (79.2{\tiny\color{green!50!black}$\uparrow$2.4}, 92.9{\tiny\color{green!50!black}$\uparrow$0.8})
           & (97.9{\tiny\color{green!50!black}$\uparrow$1.0}, 92.5{\tiny\color{green!50!black}$\uparrow$3.0})
           & (86.3{\tiny\color{green!50!black}$\uparrow$2.6}, 95.9{\tiny\color{green!50!black}$\uparrow$0.8})\\
        5  & (96.8{\tiny\color{green!50!black}$\uparrow$1.5}, \textcolor{blue}{87.7}{\tiny\color{green!50!black}$\uparrow$3.4})
           & (81.7{\tiny\color{green!50!black}$\uparrow$0.9}, 93.9{\tiny\color{green!50!black}$\uparrow$0.1})
           & (\textcolor{blue}{98.1}{\tiny\color{green!50!black}$\uparrow$1.0}, 92.7{\tiny\color{green!50!black}$\uparrow$2.9})
           & (87.2{\tiny\color{green!50!black}$\uparrow$1.3}, 96.2{\tiny\color{green!50!black}$\uparrow$0.5})\\
        7  & (\textcolor{blue}{96.7}{\tiny\color{green!50!black}$\uparrow$1.4}, 87.6{\tiny\color{green!50!black}$\uparrow$2.7})
           & (81.8{\tiny\color{green!50!black}$\uparrow$0.5}, 94.2{\tiny\color{green!50!black}$\uparrow$0.3})
           & (\textcolor{blue}{98.1}{\tiny\color{green!50!black}$\uparrow$0.8}, \textcolor{blue}{92.9}{\tiny\color{green!50!black}$\uparrow$2.9})
           & (\textcolor{blue}{87.6}{\tiny\color{green!50!black}$\uparrow$1.1}, 96.3{\tiny\color{green!50!black}$\uparrow$0.4})\\
        \rowcolor{gray!40}
        9  & (\textcolor{red}{96.9}{\tiny\color{green!50!black}$\uparrow$1.4}, \textcolor{red}{88.7}{\tiny\color{green!50!black}$\uparrow$4.3})
           & (\textcolor{blue}{82.5}{\tiny\color{green!50!black}$\uparrow$0.5}, \textcolor{red}{94.8}{\tiny\color{green!50!black}$\uparrow$0.6})
           & (\textcolor{red}{98.2}{\tiny\color{green!50!black}$\uparrow$1.0}, \textcolor{red}{93.1}{\tiny\color{green!50!black}$\uparrow$2.9})
           & (\textcolor{red}{87.7}{\tiny\color{green!50!black}$\uparrow$0.8}, \textcolor{red}{96.6}{\tiny\color{green!50!black}$\uparrow$0.5})\\
        11 & (96.6{\tiny\color{green!50!black}$\uparrow$1.2}, 87.4{\tiny\color{green!50!black}$\uparrow$3.6})
           & (\textcolor{red}{82.7}{\tiny\color{green!50!black}$\uparrow$1.0}, \textcolor{blue}{94.4}{\tiny\color{green!50!black}$\uparrow$0.2})
           & (97.9{\tiny\color{green!50!black}$\uparrow$0.8}, 92.3{\tiny\color{green!50!black}$\uparrow$2.2})
           & (87.5{\tiny\color{green!50!black}$\uparrow$2.1}, \textcolor{blue}{96.5}{\tiny\color{green!50!black}$\uparrow$1.0})\\
        \bottomrule
    \end{tabular}
\end{minipage}
\end{figure}

\begin{figure*}[t]
\begin{minipage}{0.48\textwidth}
\centering
    \vspace{-1em}
\captionof{table}{Prompt length ablation.}
      \vspace{-0.5em}
\label{table: Ablation on learnable prompt length.}
    \scriptsize
\setlength\tabcolsep{2pt}
\begin{tabular}{ccccc}
        \toprule
        \multirow{2}{*}{\makecell[c]{E}} & \multicolumn{2}{c}{Point detection} & \multicolumn{2}{c}{Multimodal detection} \\
         & Local & Global & Local & Global \\
        \midrule
        6  & (96.0{\tiny\color{green!50!black}$\uparrow$1.4}, 88.2{\tiny\color{green!50!black}$\uparrow$4.8})
           & (82.4{\tiny\color{green!50!black}$\uparrow$0.7}, 94.1{\tiny\color{purple}$\downarrow$0.1})
           & (97.9{\tiny\color{green!50!black}$\uparrow$1.4}, 92.4{\tiny\color{green!50!black}$\uparrow$2.6})
           & (87.2{\tiny\color{green!50!black}$\uparrow$0.6}, 96.3{\tiny\color{green!50!black}$\uparrow$0.3})\\
        8  & (96.2{\tiny\color{green!50!black}$\uparrow$1.0}, 88.1{\tiny\color{green!50!black}$\uparrow$4.5})
           & (82.2{\tiny\color{green!50!black}$\uparrow$0.2}, 94.6{\tiny\color{green!50!black}$\uparrow$0.4})
           & (98.1{\tiny\color{green!50!black}$\uparrow$1.3}, 92.3{\tiny\color{green!50!black}$\uparrow$2.3})
           & (87.5{\tiny\color{green!50!black}$\uparrow$0.9}, 96.3{\tiny\color{green!50!black}$\uparrow$0.5})\\
        10 & (96.5{\tiny\color{green!50!black}$\uparrow$1.2}, \textcolor{red}{88.5}{\tiny\color{green!50!black}$\uparrow$4.5})
           & (\textcolor{blue}{82.7}{\tiny\color{green!50!black}$\uparrow$0.9}, 94.5{\tiny\color{purple}$\downarrow$0.2})
           & (98.1{\tiny\color{green!50!black}$\uparrow$1.1}, 92.6{\tiny\color{green!50!black}$\uparrow$2.5})
           & (\textcolor{blue}{87.5}{\tiny\color{green!50!black}$\uparrow$1.0}, 96.3{\tiny\color{green!50!black}$\uparrow$0.2})\\
        12 & (\textcolor{red}{96.9}{\tiny\color{green!50!black}$\uparrow$1.4}, \textcolor{red}{88.7}{\tiny\color{green!50!black}$\uparrow$4.3})
           & (\textcolor{blue}{82.5}{\tiny\color{green!50!black}$\uparrow$0.5}, \textcolor{red}{94.8}{\tiny\color{green!50!black}$\uparrow$0.6})
           & (\textcolor{red}{98.2}{\tiny\color{green!50!black}$\uparrow$1.0}, \textcolor{red}{93.1}{\tiny\color{green!50!black}$\uparrow$2.9})
           & (\textcolor{red}{87.7}{\tiny\color{green!50!black}$\uparrow$0.8}, \textcolor{red}{96.6}{\tiny\color{green!50!black}$\uparrow$0.5})\\
        14 & (\textcolor{blue}{96.7}{\tiny\color{green!50!black}$\uparrow$1.3}, 88.3{\tiny\color{green!50!black}$\uparrow$4.6})
           & (82.3{\tiny\color{green!50!black}$\uparrow$0.9}, 94.4{\tiny\color{green!50!black}$\uparrow$0.3})
           & (98.0{\tiny\color{green!50!black}$\uparrow$1.1}, \textcolor{blue}{92.8}{\tiny\color{green!50!black}$\uparrow$2.7})
           & (87.3{\tiny\color{green!50!black}$\uparrow$1.8}, \textcolor{blue}{96.5}{\tiny\color{green!50!black}$\uparrow$0.9})\\
        \bottomrule
    \end{tabular}
\end{minipage}
\hfil
\begin{minipage}{0.48\textwidth}
    \centering
        \vspace{-1em}
    \captionof{table}{Density ablation.}
        \vspace{-0.5em}
    \scriptsize
 \label{table: Ablation on the point density.}
    \setlength\tabcolsep{2pt} 
    \begin{tabular}{ccccc}
\toprule
\multirow{2}{*}{\makecell[c]{Down \\ sample }} & \multicolumn{2}{c}{Point detection} & \multicolumn{2}{c}{Multimodal detection} \\
 & Local & Global & Local & Global \\
\midrule
\rowcolor{gray!40}
100\% &
(\textcolor{red}{96.9}{\tiny\color{green!50!black}$\uparrow$1.4}, \textcolor{red}{88.7}{\tiny\color{green!50!black}$\uparrow$4.3}) &
(\textcolor{blue}{82.5}{\tiny\color{green!50!black}$\uparrow$0.5}, \textcolor{blue}{94.8}{\tiny\color{green!50!black}$\uparrow$0.6}) &
(\textcolor{red}{98.2}{\tiny\color{green!50!black}$\uparrow$1.0}, \textcolor{red}{93.1}{\tiny\color{green!50!black}$\uparrow$2.9}) &
(87.7{\tiny\color{green!50!black}$\uparrow$0.8}, \textcolor{blue}{96.6}{\tiny\color{green!50!black}$\uparrow$0.5})\\
70\% &
(\textcolor{blue}{96.2}{\tiny\color{green!50!black}$\uparrow$1.3}, \textcolor{blue}{86.1}{\tiny\color{green!50!black}$\uparrow$3.5}) &
(\textcolor{red}{82.8}{\tiny\color{green!50!black}$\uparrow$1.5}, \textcolor{blue}{94.5}{\tiny\color{green!50!black}$\uparrow$0.5}) &
(\textcolor{blue}{97.9}{\tiny\color{green!50!black}$\uparrow$1.1}, \textcolor{blue}{92.6}{\tiny\color{green!50!black}$\uparrow$2.6}) &
(\textcolor{red}{88.7}{\tiny\color{green!50!black}$\uparrow$3.1}, \textcolor{red}{96.8}{\tiny\color{green!50!black}$\uparrow$1.1})\\
50\% &
(95.8{\tiny\color{green!50!black}$\uparrow$1.1}, 85.1{\tiny\color{green!50!black}$\uparrow$3.3}) &
(81.2{\tiny\color{green!50!black}$\uparrow$1.6}, 93.8{\tiny\color{green!50!black}$\uparrow$0.4}) &
(97.8{\tiny\color{green!50!black}$\uparrow$1.0}, 92.4{\tiny\color{green!50!black}$\uparrow$2.5}) &
(\textcolor{blue}{88.1}{\tiny\color{green!50!black}$\uparrow$3.2}, \textcolor{blue}{96.6}{\tiny\color{green!50!black}$\uparrow$1.1})\\
30\% &
(95.7{\tiny\color{green!50!black}$\uparrow$1.2}, 84.3{\tiny\color{green!50!black}$\uparrow$4.4}) &
(79.0{\tiny\color{green!50!black}$\uparrow$2.4}, 92.8{\tiny\color{green!50!black}$\uparrow$1.0}) &
(\textcolor{blue}{97.9}{\tiny\color{green!50!black}$\uparrow$1.5}, 92.0{\tiny\color{green!50!black}$\uparrow$2.5}) &
(86.6{\tiny\color{green!50!black}$\uparrow$3.1}, 96.2{\tiny\color{green!50!black}$\uparrow$1.1})\\
20\% &
(95.8{\tiny\color{green!50!black}$\uparrow$1.6}, 83.5{\tiny\color{green!50!black}$\uparrow$5.3}) &
(75.0{\tiny\color{green!50!black}$\uparrow$2.3}, 91.5{\tiny\color{green!50!black}$\uparrow$1.0}) &
(97.8{\tiny\color{green!50!black}$\uparrow$2.6}, 92.1{\tiny\color{green!50!black}$\uparrow$3.5}) &
(85.3{\tiny\color{green!50!black}$\uparrow$3.3}, 95.7{\tiny\color{green!50!black}$\uparrow$1.1})\\
\bottomrule
\end{tabular}
	\end{minipage}
\end{figure*}

\begin{figure*}
\begin{minipage}{0.48\textwidth}
    \centering
    \vspace{-1em}
    \captionof{table}{Lighting ablation.}%
    \vspace{-0.5em}
    \label{tabel: Visualization with different rendering lighting.}
    \scriptsize
    \setlength\tabcolsep{2pt}
    \begin{tabular}{ccccc}
\toprule
\multirow{2}{*}{Lighting} & \multicolumn{2}{c}{Point detection} & \multicolumn{2}{c}{Multimodal detection} \\
 & Local & Global & Local & Global \\
\midrule
++ &
(\textcolor{red}{96.9}{\tiny\color{green!50!black}$\uparrow$1.5}, \textcolor{blue}{88.3}{\tiny\color{green!50!black}$\uparrow$4.6}) &
(82.1{\tiny\color{green!50!black}$\uparrow$0.1}, \textcolor{blue}{94.8}{\tiny\color{green!50!black}$\uparrow$0.5}) &
(\textcolor{red}{98.3}{\tiny\color{green!50!black}$\uparrow$1.1}, 92.8{\tiny\color{green!50!black}$\uparrow$2.1}) &
(\textcolor{blue}{87.6}{\tiny\color{green!50!black}$\uparrow$1.9}, \textcolor{blue}{96.4}{\tiny\color{green!50!black}$\uparrow$0.6})\\
+ &
(\textcolor{red}{96.9}{\tiny\color{green!50!black}$\uparrow$1.5}, \textcolor{blue}{88.3}{\tiny\color{green!50!black}$\uparrow$4.5}) &
(\textcolor{blue}{82.0}{\tiny\color{purple}$\downarrow$0.4}, \textcolor{red}{95.1}{\tiny\color{green!50!black}$\uparrow$0.5}) &
(\textcolor{red}{98.3}{\tiny\color{green!50!black}$\uparrow$1.2}, \textcolor{blue}{93.0}{\tiny\color{green!50!black}$\uparrow$2.5}) &
(87.5{\tiny\color{green!50!black}$\uparrow$1.4}, \textcolor{blue}{96.4}{\tiny\color{green!50!black}$\uparrow$0.5})\\
\rowcolor{gray!40}
Ori &
(\textcolor{red}{96.9}{\tiny\color{green!50!black}$\uparrow$1.4}, \textcolor{red}{88.7}{\tiny\color{green!50!black}$\uparrow$4.3}) &
(\textcolor{blue}{82.5}{\tiny\color{green!50!black}$\uparrow$0.5}, \textcolor{blue}{94.8}{\tiny\color{green!50!black}$\uparrow$0.6}) &
(\textcolor{red}{98.2}{\tiny\color{green!50!black}$\uparrow$1.0}, \textcolor{red}{93.1}{\tiny\color{green!50!black}$\uparrow$2.9}) &
(\textcolor{red}{87.7}{\tiny\color{green!50!black}$\uparrow$0.8}, \textcolor{red}{96.6}{\tiny\color{green!50!black}$\uparrow$0.5})\\
- &
(\textcolor{blue}{96.7}{\tiny\color{green!50!black}$\uparrow$1.4}, 87.9{\tiny\color{green!50!black}$\uparrow$4.0}) &
(81.8{\tiny\color{purple}$\downarrow$0.6}, \textcolor{blue}{94.8}{\tiny\color{green!50!black}$\uparrow$0.3}) &
(\textcolor{blue}{98.2}{\tiny\color{green!50!black}$\uparrow$1.1}, 92.6{\tiny\color{green!50!black}$\uparrow$2.0}) &
(86.9{\tiny\color{green!50!black}$\uparrow$0.5}, 96.2{\tiny\color{green!50!black}$\uparrow$0.3})\\
– &
(\textcolor{blue}{96.7}{\tiny\color{green!50!black}$\uparrow$1.4}, 87.7{\tiny\color{green!50!black}$\uparrow$4.3}) &
(81.5{\tiny\color{purple}$\downarrow$0.4}, \textcolor{blue}{94.8}{\tiny\color{green!50!black}$\uparrow$0.5}) &
(\textcolor{blue}{98.2}{\tiny\color{green!50!black}$\uparrow$1.1}, 92.7{\tiny\color{green!50!black}$\uparrow$2.2}) &
(87.0{\tiny\color{green!50!black}$\uparrow$0.9}, 96.2{\tiny\color{green!50!black}$\uparrow$0.4})\\
\bottomrule
\end{tabular}
\end{minipage}
\hfil
\begin{minipage}{0.48\textwidth}
    \centering
    \vspace{-1em}
    \captionof{table}{ \textcolor{orange}{Occlusion ablation.}}%
    \vspace{-0.5em}
    \label{table Analysis on the point occlusion.}
    \scriptsize
    \setlength\tabcolsep{2pt}
    \begin{tabular}{ccccc}
\toprule
\multirow{2}{*}{Method} & \multicolumn{2}{c}{Point detection} & \multicolumn{2}{c}{Multimodal detection} \\
 & Local & Global & Local & Global \\
\midrule
\rowcolor{gray!40}
Ori &
(\textcolor{red}{96.9}{\tiny\color{green!50!black}$\uparrow$1.4}, \textcolor{red}{88.7}{\tiny\color{green!50!black}$\uparrow$4.3}) &
(\textcolor{red}{82.5}{\tiny\color{green!50!black}$\uparrow$0.5}, \textcolor{red}{94.8}{\tiny\color{green!50!black}$\uparrow$0.6}) &
(\textcolor{red}{98.2}{\tiny\color{green!50!black}$\uparrow$1.0}, \textcolor{red}{93.1}{\tiny\color{green!50!black}$\uparrow$2.9}) &
(\textcolor{red}{87.7}{\tiny\color{green!50!black}$\uparrow$0.8}, \textcolor{red}{96.6}{\tiny\color{green!50!black}$\uparrow$0.5})\\
\rowcolor{gray!10}
Occlu. &
(\textcolor{blue}{95.4}{\tiny\color{green!50!black}$\uparrow$1.1}, \textcolor{blue}{83.9}{\tiny\color{green!50!black}$\uparrow$3.1}) &
(\textcolor{blue}{73.6}{\tiny\color{green!50!black}$\uparrow$0.3}, \textcolor{blue}{90.6}{\tiny\color{green!50!black}$\uparrow$0.0}) &
(\textcolor{blue}{97.5}{\tiny\color{green!50!black}$\uparrow$0.8}, \textcolor{blue}{90.7}{\tiny\color{green!50!black}$\uparrow$1.2}) &
(\textcolor{blue}{83.9}{\tiny\color{green!50!black}$\uparrow$0.9}, \textcolor{blue}{95.2}{\tiny\color{green!50!black}$\uparrow$0.4})\\
\bottomrule
\end{tabular}
\end{minipage}
\end{figure*}

\begin{figure*}[h]
  \begin{minipage}{0.48\textwidth}
        \centering
\vspace{-1em}
 \captionof{table}{ \textcolor{orange}{Quality ablation.}}
 \vspace{-0.5em}
 \label{table: Ablation on the rendering quality.}
 \scriptsize
    \setlength\tabcolsep{2pt} 
    \begin{tabular}{ccccc}
        \toprule
         \multirow{2}{*}{\makecell[c]{Blur  \\ sample }} & \multicolumn{2}{c}{Point detection} & \multicolumn{2}{c}{Multimodal detection} \\
         & Local & Global & Local & Global \\
        \midrule
        \rowcolor{gray!40}
        0
        & (\textcolor{red}{96.9}{\tiny\color{green!50!black}$\uparrow$1.4}, \textcolor{red}{88.7}{\tiny\color{green!50!black}$\uparrow$4.3})
        & (\textcolor{blue}{82.5}{\tiny\color{green!50!black}$\uparrow$0.5}, \textcolor{red}{94.8}{\tiny\color{green!50!black}$\uparrow$0.6})
        & (\textcolor{red}{98.2}{\tiny\color{green!50!black}$\uparrow$1.0}, \textcolor{red}{93.1}{\tiny\color{green!50!black}$\uparrow$2.9})
        & (\textcolor{red}{87.7}{\tiny\color{green!50!black}$\uparrow$0.8}, \textcolor{blue}{96.6}{\tiny\color{green!50!black}$\uparrow$0.5})\\
        1
        & (\textcolor{blue}{96.8}{\tiny\color{green!50!black}$\uparrow$1.6}, \textcolor{blue}{88.1}{\tiny\color{green!50!black}$\uparrow$5.5})
        & (\textcolor{blue}{81.3}{\tiny\color{green!50!black}$\uparrow$1.2}, \textcolor{blue}{93.9}{\tiny\color{green!50!black}$\uparrow$0.4})
        & (\textcolor{blue}{98.1}{\tiny\color{green!50!black}$\uparrow$1.1}, \textcolor{blue}{93.0}{\tiny\color{green!50!black}$\uparrow$3.0})
        & (\textcolor{blue}{87.0}{\tiny\color{green!50!black}$\uparrow$1.5}, \textcolor{blue}{96.2}{\tiny\color{green!50!black}$\uparrow$0.6})\\
        5
        & (96.6{\tiny\color{green!50!black}$\uparrow$1.5}, 87.4{\tiny\color{green!50!black}$\uparrow$5.1})
        & (77.7{\tiny\color{purple}$\downarrow$0.5}, 92.8{\tiny\color{green!50!black}$\uparrow$0.3})
        & (98.0{\tiny\color{green!50!black}$\uparrow$1.2}, 92.2{\tiny\color{green!50!black}$\uparrow$2.5})
        & (85.4{\tiny\color{green!50!black}$\uparrow$1.8}, 95.7{\tiny\color{green!50!black}$\uparrow$0.6})\\
        9
        & (96.7{\tiny\color{green!50!black}$\uparrow$1.6}, 87.1{\tiny\color{green!50!black}$\uparrow$4.8})
        & (77.5{\tiny\color{purple}$\downarrow$0.1}, 92.3{\tiny\color{green!50!black}$\uparrow$0.1})
        & (97.8{\tiny\color{green!50!black}$\uparrow$1.3}, 92.0{\tiny\color{green!50!black}$\uparrow$2.7})
        & (85.1{\tiny\color{green!50!black}$\uparrow$1.9}, 95.5{\tiny\color{green!50!black}$\uparrow$0.5})\\
        \bottomrule
    \end{tabular}
    \end{minipage} 
    \hfil
\begin{minipage}{0.48\textwidth}
        \centering
\vspace{-0.5em}
 \captionof{table}{ \textcolor{orange}{Angle ablation.}}
 \vspace{-1em}
 \label{table Analysis on the rendering angle.}
 \scriptsize
    \setlength\tabcolsep{2pt} 
\begin{tabular}{ccccc}
\toprule
\multirow{2}{*}{\makecell[c]{Angle \\ shift}} & \multicolumn{2}{c}{Point detection} & \multicolumn{2}{c}{Multimodal detection} \\
 & Local & Global & Local & Global \\
\midrule
\rowcolor{gray!40}
0 &
(\textcolor{red}{96.9}{\tiny\color{green!50!black}$\uparrow$1.4}, \textcolor{red}{88.7}{\tiny\color{green!50!black}$\uparrow$4.3}) &
(82.5{\tiny\color{green!50!black}$\uparrow$0.5}, 94.8{\tiny\color{green!50!black}$\uparrow$0.6}) &
(\textcolor{red}{98.2}{\tiny\color{green!50!black}$\uparrow$1.0}, \textcolor{red}{93.1}{\tiny\color{green!50!black}$\uparrow$2.9}) &
(\textcolor{blue}{87.7}{\tiny\color{green!50!black}$\uparrow$0.8}, \textcolor{red}{96.6}{\tiny\color{green!50!black}$\uparrow$0.5})\\
$\tfrac{1}{20}\pi$ &
(96.6{\tiny\color{green!50!black}$\uparrow$1.2}, 87.8{\tiny\color{green!50!black}$\uparrow$3.9}) &
(83.2{\tiny\color{green!50!black}$\uparrow$0.6}, 94.9{\tiny\color{green!50!black}$\uparrow$0.3}) &
(\textcolor{blue}{98.1}{\tiny\color{green!50!black}$\uparrow$1.0}, 92.6{\tiny\color{green!50!black}$\uparrow$1.9}) &
(87.2{\tiny\color{green!50!black}$\uparrow$0.5}, \textcolor{red}{96.6}{\tiny\color{green!50!black}$\uparrow$0.6})\\
$\tfrac{2}{20}\pi$ &
(\textcolor{blue}{96.7}{\tiny\color{green!50!black}$\uparrow$1.3}, \textcolor{blue}{88.0}{\tiny\color{green!50!black}$\uparrow$3.9}) &
(\textcolor{blue}{83.2}{\tiny\color{green!50!black}$\uparrow$1.1}, 94.9{\tiny\color{green!50!black}$\uparrow$0.5}) &
(\textcolor{blue}{98.1}{\tiny\color{green!50!black}$\uparrow$1.0}, \textcolor{blue}{92.7}{\tiny\color{green!50!black}$\uparrow$2.0}) &
(\textcolor{red}{87.9}{\tiny\color{green!50!black}$\uparrow$1.5}, \textcolor{blue}{96.5}{\tiny\color{green!50!black}$\uparrow$0.6})\\
$\tfrac{3}{20}\pi$ &
(\textcolor{blue}{96.7}{\tiny\color{green!50!black}$\uparrow$1.3}, 87.9{\tiny\color{green!50!black}$\uparrow$3.6}) &
(\textcolor{red}{84.1}{\tiny\color{green!50!black}$\uparrow$1.5}, \textcolor{red}{95.2}{\tiny\color{green!50!black}$\uparrow$0.6}) &
(\textcolor{blue}{98.1}{\tiny\color{green!50!black}$\uparrow$1.0}, 92.6{\tiny\color{green!50!black}$\uparrow$1.9}) &
(87.2{\tiny\color{green!50!black}$\uparrow$0.8}, \textcolor{red}{96.6}{\tiny\color{green!50!black}$\uparrow$0.7})\\
\bottomrule
\end{tabular}
    \end{minipage} 
\end{figure*}

 \textcolor{orange}{
\textbf{Ablation on aggregation strategy.}
By default, we adopt the mean aggregation strategy to obtain explicit point representations. We also investigate two alternative strategies: \textbf{max strategy}, which selects the maximum pixel representation across different views as the corresponding explicit point representation; \textbf{network-parametric strategy}, which employs an MLP followed by a softmax layer to compute aggregation weights of the pixel representations. As shown in Table~\ref{table Ablation on aggregation strategy.}, PointAD+ is not restricted to a specific aggregation scheme but generalizes across different strategies. We adopt the mean strategy as it provides a favorable trade-off between performance and computational overhead.
}

 \textcolor{orange}{
\textbf{Ablation on $\alpha$.}
The coefficient $\alpha$ balances the feature fusion between the F and S stages. We investigate the effect of the coefficient $\alpha$ on the model's performance. We set $\alpha \in \{0, 0.1, 0.3, 0.5, 0.7,0.9, 1\}$. As shown in Figure~\ref{fig: hyperparameter Ablation}, increasing $\alpha$ initially improves segmentation performance. However, as $\alpha$ continues to increase, performance begins to decline. Empirically, $\alpha = 0.5$ achieves a good balance between point and multimodal segmentation performance.
}

\textbf{Ablation on view number.} PointAD+ interprets point clouds through 2D renderings, where the number of rendering views directly influences the extent of 3D information captured. As shown in Table~\ref{table: Ablation on the view number.}, an excessive number of views introduces redundancy and noise, hindering accurate point representation. Conversely, an insufficient number of views results in information loss. Therefore, selecting an optimal number of views enhances point cloud understanding by effectively leveraging informative perspectives.

\textbf{Ablation on length of learnable prompts.} Here, we examine the impact of the length of learnable text prompt templates. As shown in Table~\ref{table: Ablation on learnable prompt length.}, increasing the length of word embeddings enhances PointAD+'s capacity to incorporate both 3D and 2D anomaly semantics, contributing to the performance gain. However, further increasing the length (e.g., from 12 to 14) does not yield additional performance gains and instead results in a declining trend. Excessively long prompts may introduce noise, which disrupts the detection process. This demonstrates that both excessive and insufficient lengths of learnable word embeddings can degrade performance. In our study, an optimal prompt length of 12 enables PointAD+ to achieve comprehensive performance in both 3D and multimodal 3D anomaly detection.

\begin{figure*}
\centering
    \subfigure[Visualization on varied densities downsampled from original point clouds.]
    {\includegraphics[width=0.46\textwidth]
    {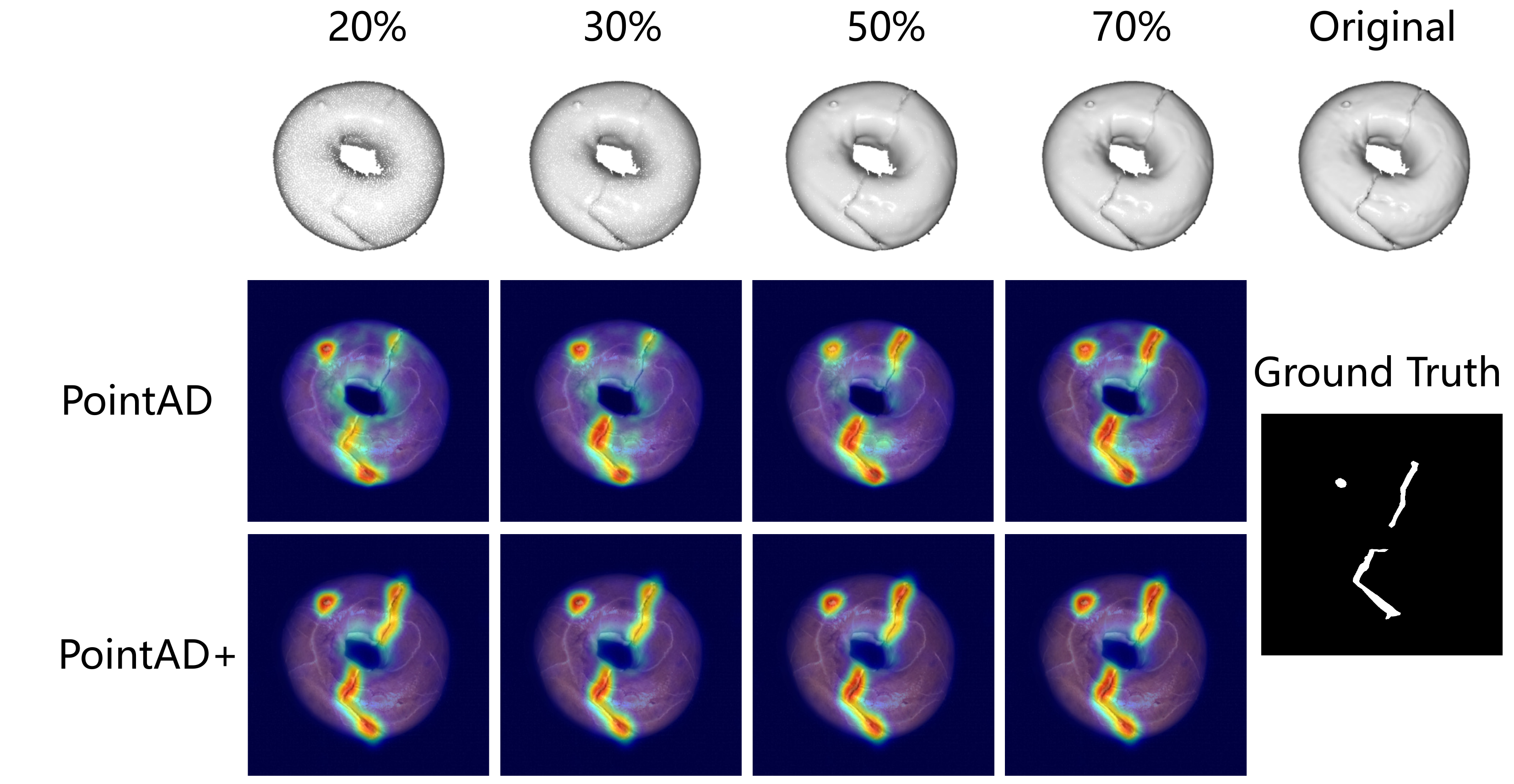}
    \label{fig: Visualization with different resolutions}}
    \subfigure[Visualization on different rendering lighting.]
    {\includegraphics[width=0.46\textwidth]{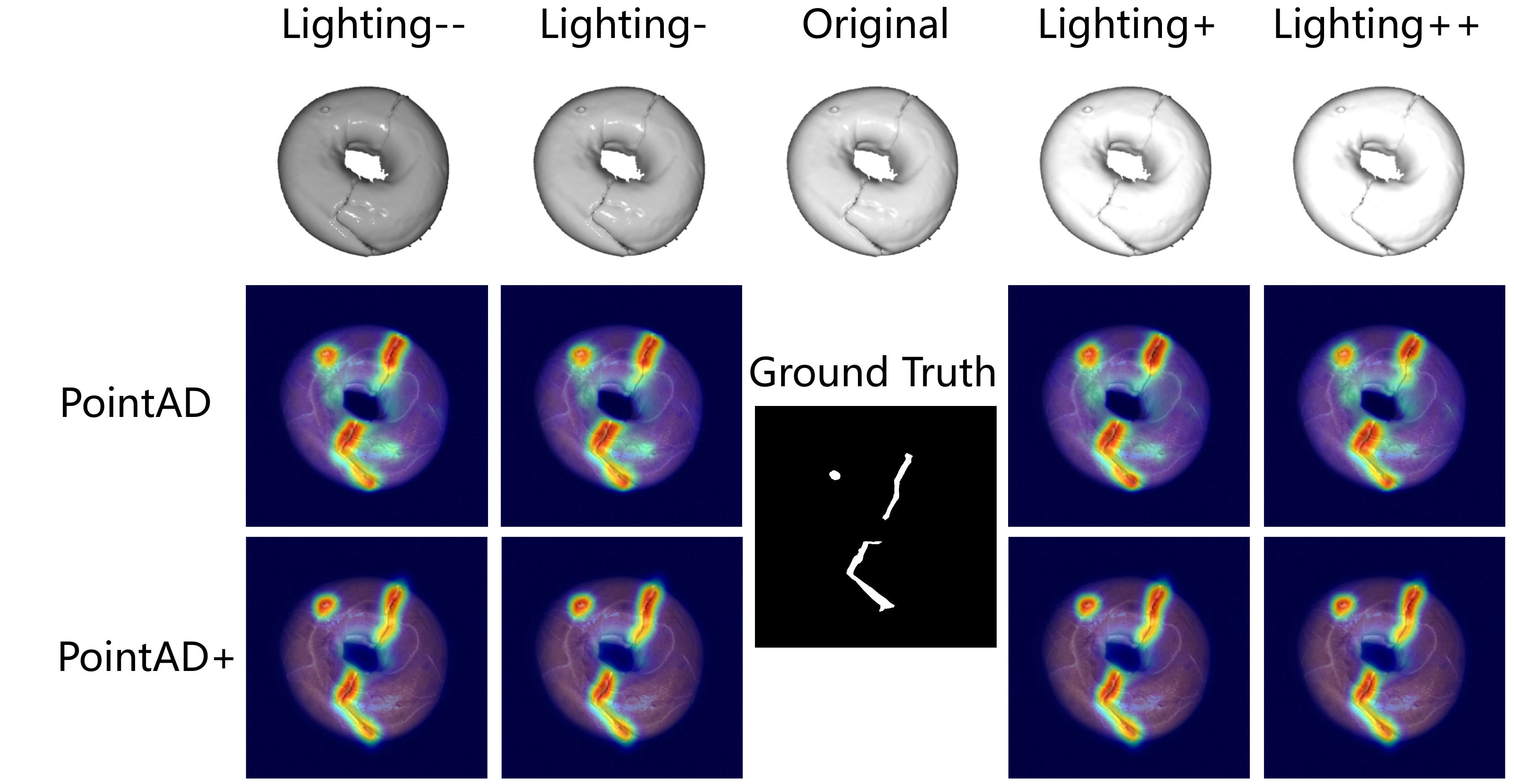}
\label{fig: Visualization with different rendering lighting.}}
\end{figure*}

 \textcolor{orange}{
\subsection{Ablation on the varied rendering conditions.}
PointAD+ interprets point clouds based on the rendering representations through CLIP. Here, we investigate the effect of the rendering condition, including point density in Table~\ref{table: Ablation on the point density.}, lighting ablation in Table~\ref{tabel: Visualization with different rendering lighting.}, point occlusion in Table~\ref{table Analysis on the point occlusion.}, rendering quality in Table~\ref{table: Ablation on the rendering quality.}, and rendering angle in Table~\ref{table Analysis on the rendering angle.}. 
}

\textbf{Ablation on input density.} We investigate the impact of input point density on the performance of PointAD+ and PointAD. To create low-resolution point clouds, we downsample the original high-resolution point clouds using Farthest Point Sampling (FPS) with various sampling ratios. This generates corresponding low-resolution datasets for evaluating PointAD+. Visualizations of these datasets are provided in Figure~\ref{fig: Visualization with different resolutions}. We train PointAD+ on the original dataset and evaluate its performance on point clouds with varied resolutions. As shown in Table~\ref{table: Ablation on the point density.}, PointAD+ demonstrates robust detection capabilities across a wide range of resolutions, including downsampling ratios of 20\%, 30\%, 50\%, and 70\%. Notably, PointAD+ achieves SOTA performance even at 20\% resolution, indicating that it is highly adaptable to point clouds with varying resolutions.

\textbf{Ablation on rendering lighting.} To examine the impact of rendering lighting, we adjusted the lighting settings used to render point clouds in Open3D, generating a series of datasets with a range of lighting intensities. Specifically, we applied both stronger and weaker lighting compared to the original dataset, covering a broad range of lighting conditions. We denote stronger and the strongest lighting levels as "+" and "++," respectively, and weaker and the weakest lighting levels as "-" and "--". Figure~\ref{fig: Visualization with different rendering lighting.} presents the visualizations of the resulting samples on MVTec3D-AD, where PointAD+ was trained on the original dataset and tested on these lighting-variant datasets. As shown in Table~\ref{tabel: Visualization with different rendering lighting.}, PointAD+ consistently detects anomalies despite significant variations in rendering lighting, demonstrating its robustness to such discrepancies.

 \textcolor{orange}{
\textbf{Ablation on rendering angles.}
To further examine the effect of the absolute angle, we adjust the rendering orientations while preserving the relative angular discrepancy. In our setup, the baseline adjacent angle discrepancy is $\tfrac{1}{5}\pi$. This gap is subdivided into four intervals, and we apply angular shifts of $\tfrac{1}{20}\pi$, $\tfrac{2}{20}\pi$, and $\tfrac{3}{20}\pi$ to evaluate the influence of different orientation changes. As reported in Table~\ref{table Analysis on the rendering angle.}, PointAD+ delivers stable performance under these variations and maintains the superiority of PointAD. This confirms its robustness to unseen rendering angles.
}

 \textcolor{orange}{
\textbf{Ablation on rendering quality.}
 In our experiments, we relied on the Open3D Library to generate point cloud renderings, though it does not provide an API for explicitly controlling rendering quality. Here, we apply a Gaussian blur to the 2D renderings to simulate the noise level. We evaluated PointAD on MVTec3D-AD under blur levels $\sigma \in \{1, 5, 9\}$. As summarized in Table~\ref{table: Ablation on the rendering quality.}, the detection performance of PointAD+ decreases as the rendering quality degrades (i.e., with larger $\sigma$ values). Nevertheless, even under heavy blur ($\sigma=9$), the model maintains acceptable accuracy and consistently outperforms baselines that rely on high-quality renderings.
}

 \textcolor{orange}{
\textbf{Ablation on point occlusion.}
We next evaluated detection performance under occluded point clouds by removing points invisible from a given rendering angle and projecting the remaining ones. In some cases, abnormal regions were completely hidden, converting anomalies into normal instances. To avoid this effect, we selected angles that preserved visibility of anomalous regions. Table~\ref{table Analysis on the point occlusion.} shows that PointAD+ exhibits performance degradation when tested on the occluded point clouds. We attribute this degradation to two factors: first, occluded point clouds may lose part of the anomaly semantics, making anomalies harder to recognize; second, occlusion can create artificial sinkholes on the surface, which may cause PointAD+ to mistakenly identify these regions as hole anomalies. Although PointAD+ outperforms PointAD, its performance gain on the occluded dataset is smaller than on the original dataset. We analyze that PointAD+ leverages intrinsic point relations. When points in certain regions are completely removed, these relations are disrupted and hinder geometric modeling.
}

\section{Conclusion}
This paper addresses the challenging yet underexplored tasks of ZS 3D and multimodal 3D anomaly detection.  We begin by analyzing point abnormality from two perspectives: explicit 3D abnormality, which aims to uncover unusual spatial relations, and implicit 3D abnormality, which interprets 3D anomalies based on rendering pixel anomalies. PointAD introduces hybrid representation learning to capture implicit 3D abnormality. Further, we propose hierarchical representation learning to jointly learn explicit 3D abnormality on the geometric layer and implicit 3D abnormality on the rendering layer. This results in PointAD+, which effectively recognizes generic 3D normality and abnormality across diverse objects and seamlessly integrates RGB information for ZS multimodal 3D detection. Quantitative results demonstrate the superior ZS detection performance of our models, whether in single-modality or multi-modality scenarios. In addition, we provide an analysis of how our proposed modules function and complement each other. Extensive ablation studies on hyperparameters demonstrate their robustness across varying parameter settings. 

Although PointAD+ and PointAD have demonstrated superior performance, there remain several directions for further exploration. One promising avenue is the development of an additional point encoder to incorporate generalized geometric information into explicit point representations, further bridging the point relation gap between seen and unseen point clouds. Another potential direction is to explore a fine-grained filtering mechanism for selecting high-quality 2D renderings, particularly to better highlight anomalies. \textbf{Code will be made available once the paper is accepted.}

\bibliography{neurips_2024}
\bibliographystyle{IEEEtran}

\begin{IEEEbiography}[{\includegraphics[width=1in,height=1.25in,clip,keepaspectratio]{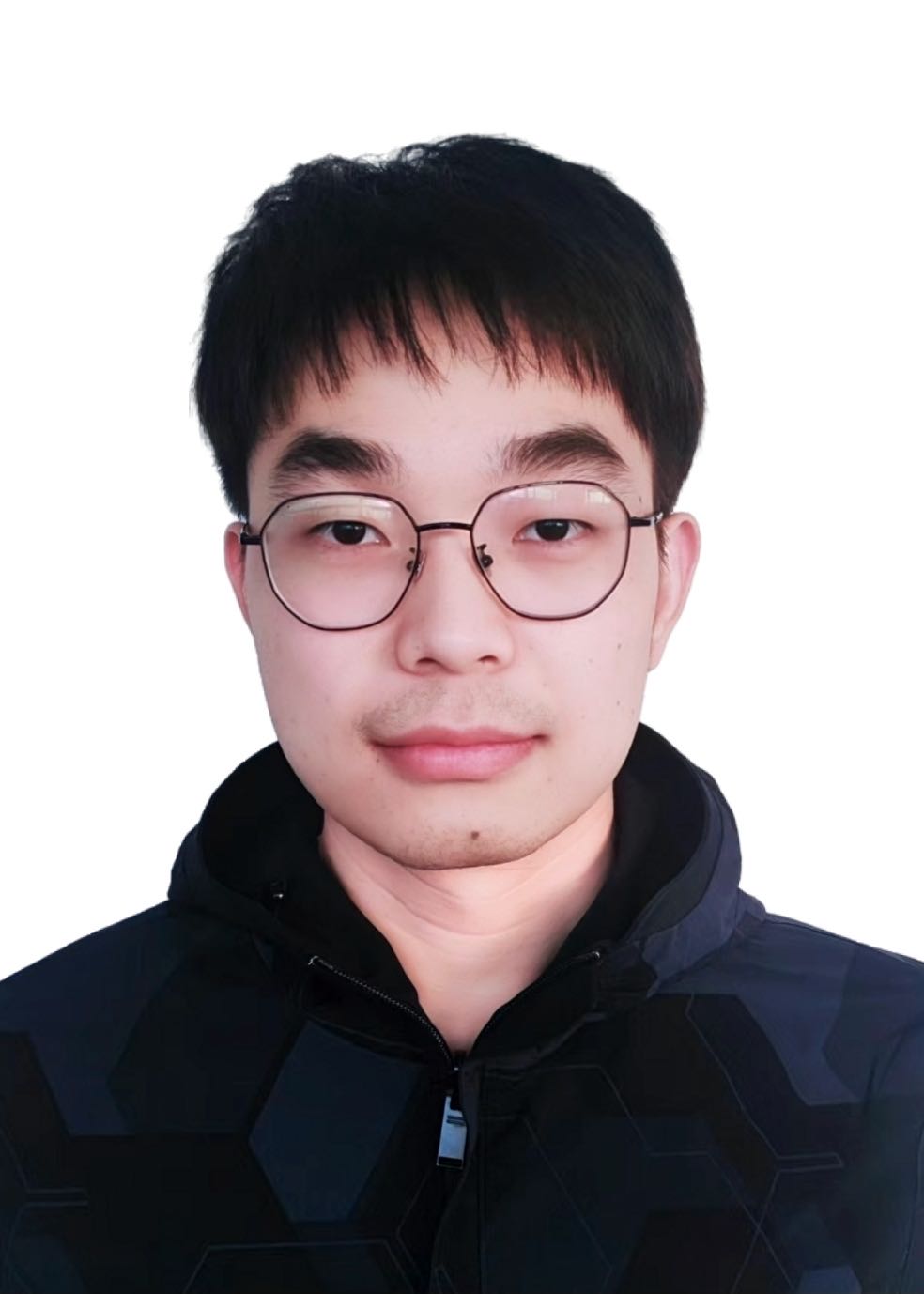}}]{Qihang Zhou} received
the B.S. degree from China University of Geosciences, Wuhan, China, in 2020. He is currently pursuing the Ph.D. degree in control science
and engineering with the College of Control Science
and Engineering, Zhejiang University, Hangzhou,
China. His research interests include Vision-language models, anomaly detection, medical analysis, and ZS learning.
\end{IEEEbiography}

\begin{IEEEbiography}[{\includegraphics[width=1in,height=1.25in,clip,keepaspectratio]{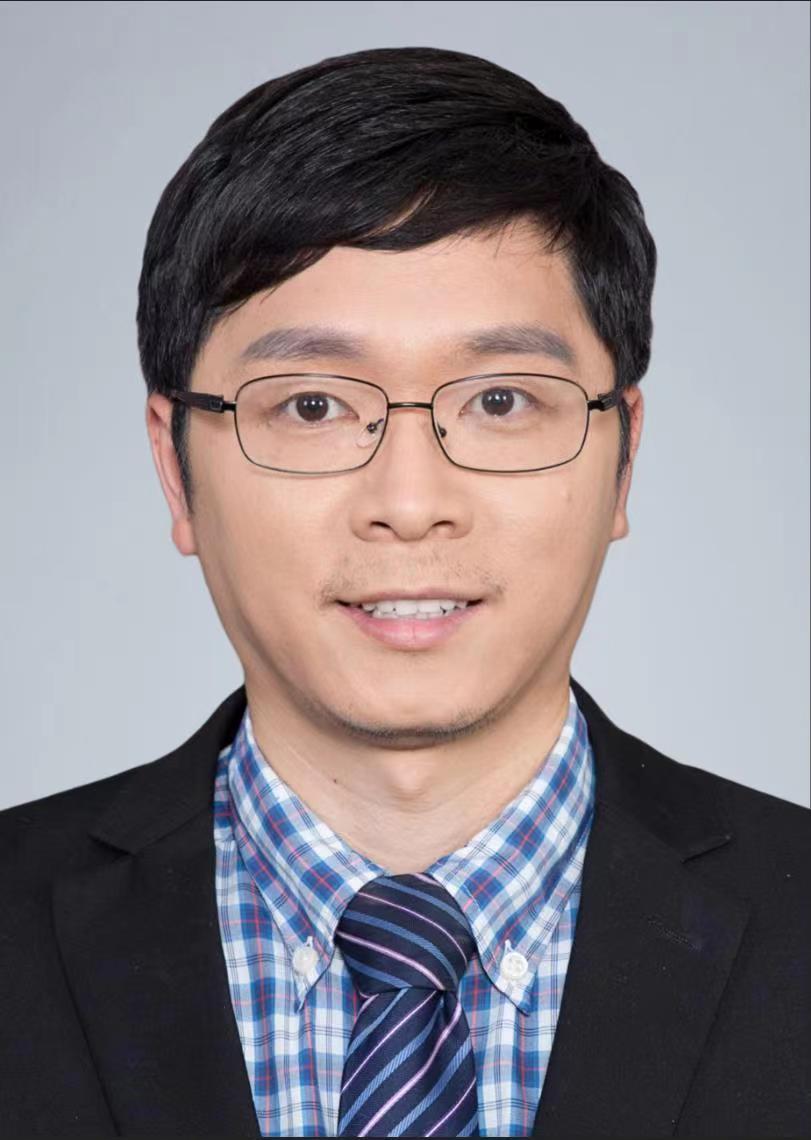}}]{Shibo He}
(Senior Member, IEEE) received the Ph.D. degree
in control science and engineering from Zhejiang
University, Hangzhou, China, in 2012. He is currently
a Professor with Zhejiang University. He
was an Associate Research Scientist from March
2014 to May 2014, and a Postdoctoral Scholar from
May 2012 to February 2014, with Arizona State
University, Tempe, AZ, USA. From November 2010
to November 2011, he was a Visiting Scholar with
the University of Waterloo, Waterloo, ON, Canada.
His research interests include Internet of Things,
crowdsensing, big data analysis, etc. Prof. He serves on the editorial board
for the IEEE TRANSACTIONS ON VEHICULAR TECHNOLOGY, Springer
Peer-to-Peer Networking and Application and KSII Transactions on Internet
and Information Systems, and is a Guest Editor for Elsevier Computer
Communications and Hindawi International Journal of Distributed Sensor
Networks. He was a General Co-chair for iSCI 2022, Symposium Co-Chair for the IEEE/CIC ICCC 2022, IEEE GlobeCom 2020
and the IEEE ICC 2017, TPC Co-Chair for i-Span 2018, a Finance and
Registration chair for ACM MobiHoc 2015, a TPC Co-Chair for the IEEE
ScalCom 2014, a TPC Vice Co-Chair for ANT 2013 and 2014, a Track Co-Chair
for the Pervasive Algorithms, Protocols, and Networks of EUSPN 2013, a
Web Co-Chair for the IEEE MASS 2013, and a Publicity Co-Chair of IEEE
WiSARN 2010, and FCN 2014.
\end{IEEEbiography}

\begin{IEEEbiography}[{\includegraphics[width=1in,height=1.25in,clip,keepaspectratio]{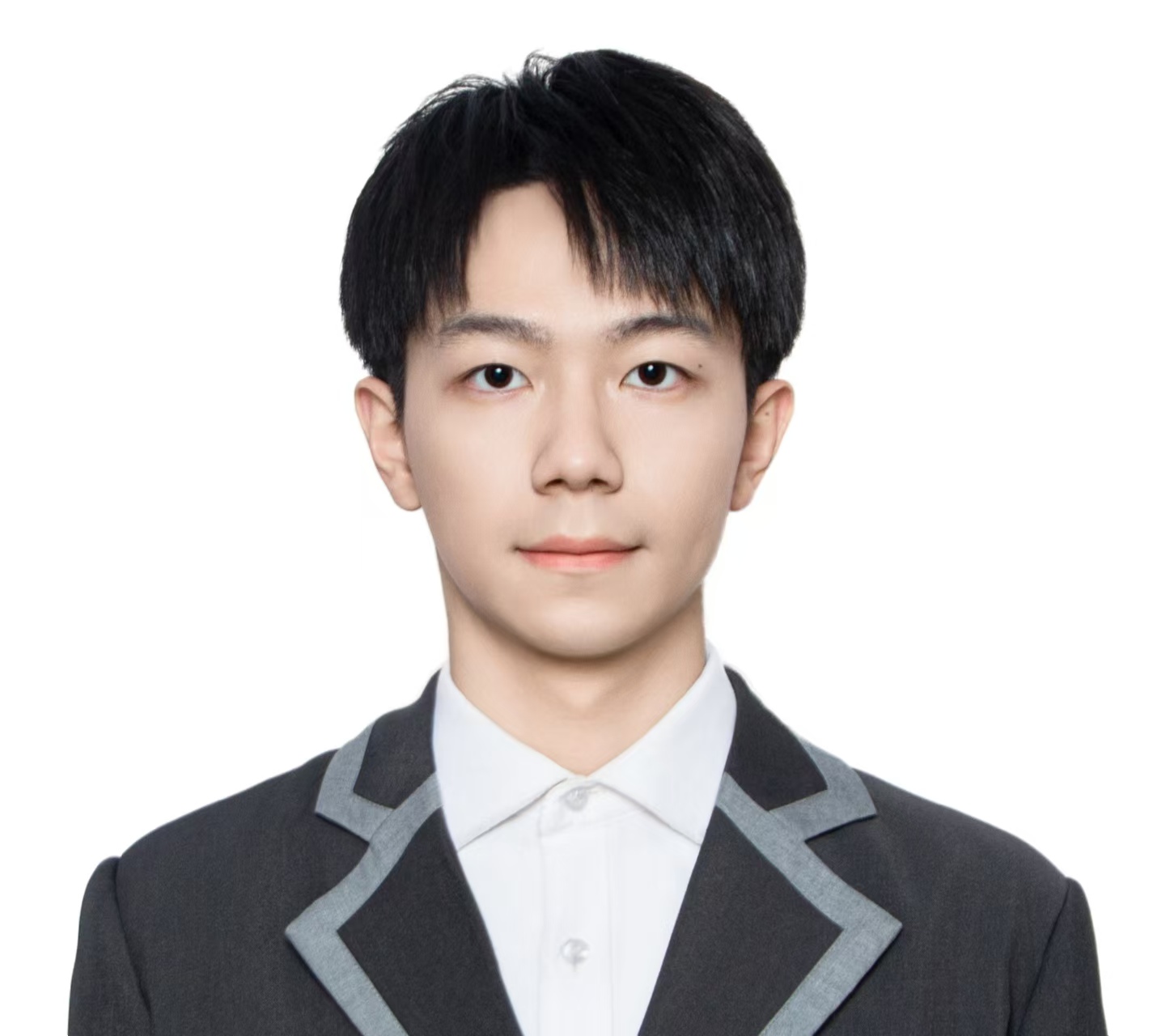}}]{Jiangtao Yan} received the B.S. degree from Huazhong University of Science and Technology, Wuhan, China, in 2022. He is currently pursuing the M.S. degree in control science and engineering with the College of Control Science and Engineering, Zhejiang University, Hangzhou, China. His research interests include anomaly detection and Vision-language models.

\end{IEEEbiography}

\begin{IEEEbiography}[{\includegraphics[width=1in,height=1.25in,clip,keepaspectratio]{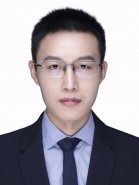}}]{Wenchao Meng}
(Senior Member, IEEE) received
the Ph.D. degree in control science and engineering from Zhejiang University, Hangzhou, China, in 2015, where he is currently with the College
of Control Science and Engineering. His current research interests include adaptive intelligent
control, cyber–physical systems, renewable energy
systems, and smart grids.
\end{IEEEbiography}

\begin{IEEEbiography}[{\includegraphics[width=1in,height=1.25in,clip,keepaspectratio]{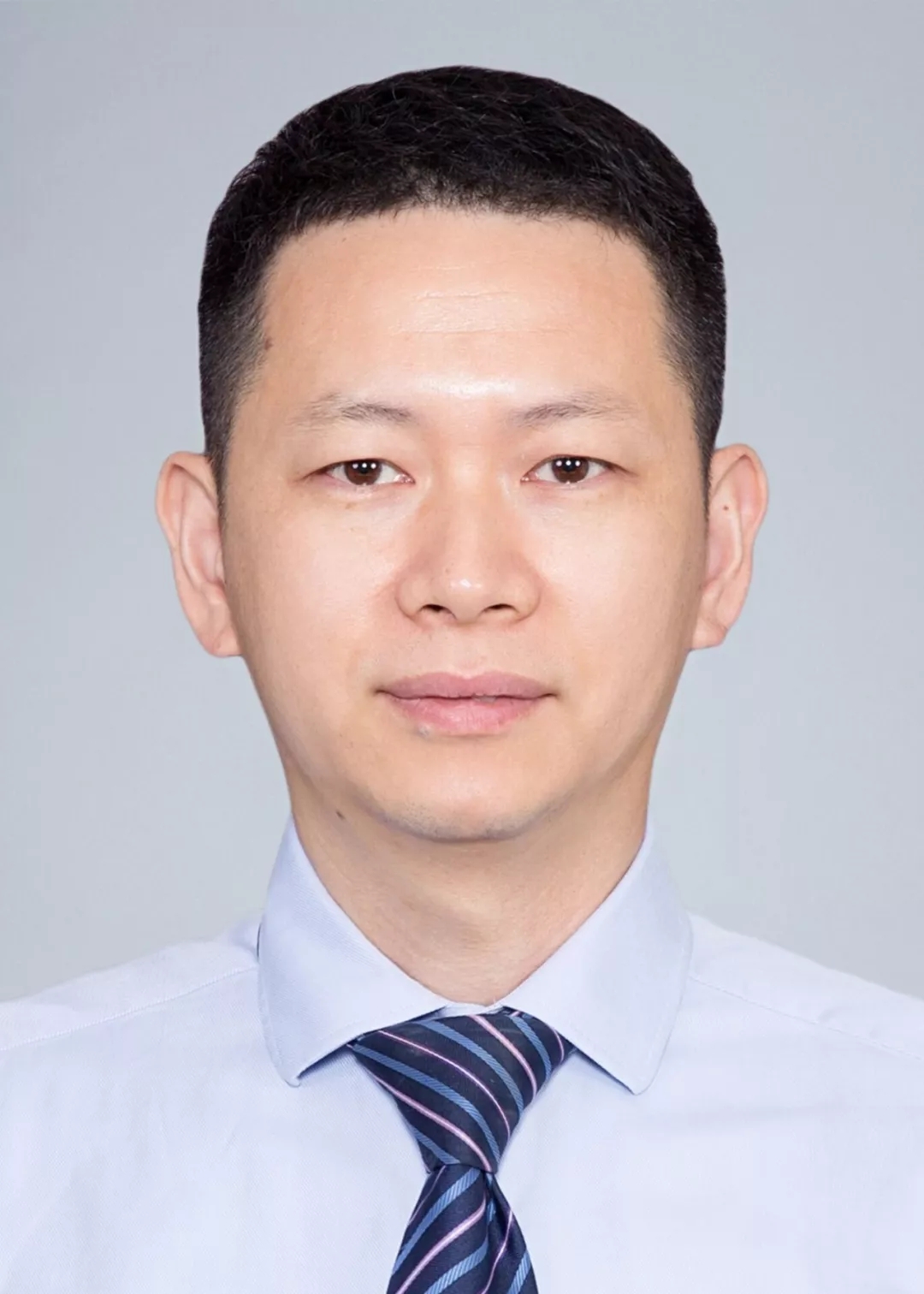}}]{Jiming Chen}
(Fellow, IEEE) received the PhD degree in control science and engineering from Zhejiang University, Hangzhou, China, in 2005. He is currently a professor with the Department of Control Science and Engineering, the vice dean of the Faculty of Information Technology, Zhejiang University. His research interests include IoT, networked control, wireless networks. He serves on the editorial boards of multiple IEEE Transactions, and the general co-chairs for IEEE RTCSA’19, IEEE Datacom’19 and IEEE PST’20. He was a recipient of the 7th IEEE ComSoc Asia/Pacific Outstanding Paper Award, the JSPS Invitation Fellowship, and the IEEE ComSoc AP Outstanding Young Researcher Award. He is an IEEE VTS distinguished lecturer. He is a fellow of the CAA.
\end{IEEEbiography}

\end{document}